\pdfoutput=1
\PassOptionsToPackage{dvipsnames}{xcolor}
\documentclass[11pt]{article}

\usepackage[final]{acl}

\usepackage{times}
\usepackage{latexsym}
\usepackage{float}
\usepackage{todonotes}
\usepackage[most]{tcolorbox}
\usepackage{booktabs}
\usepackage{amsmath} 
\usepackage{titlefoot}
\usepackage{pifont} 
\usepackage{marvosym}
\usepackage{authblk}

\usepackage[T1]{fontenc}

\usepackage[utf8]{inputenc}
\usepackage[most]{tcolorbox}
\usepackage{siunitx}
\usepackage{microtype}
\usepackage{amsfonts} 
\usepackage{pifont}
\usepackage{cancel}
\usepackage{soul}
\usepackage{bm}
\usepackage{algorithm}
\usepackage{algorithmic}
\usepackage{booktabs}
\usepackage{xspace}
\usepackage{fontawesome}
\usepackage{mathrsfs}
\usepackage{multirow}
\usepackage{amsmath}
\usepackage{colortbl}
\usepackage{subfig}
\usepackage{graphicx}
\usepackage{tabularx}
\usepackage{arydshln}
\usepackage{makecell}
\usepackage{CJKutf8}
\usepackage{adjustbox}

\usepackage[dvipsnames]{xcolor}
\newcommand{\stdvu}[1]{\scriptsize{\color{darkgray}(#1)}}

\newcolumntype{L}[1]{>{\raggedright\arraybackslash}p{#1}}
\newcolumntype{C}[1]{>{\centering\arraybackslash}m{#1}}
\newcolumntype{R}[1]{>{\raggedleft\arraybackslash}p{#1}}

\title{Beyond Survival: Evaluating LLMs in Social Deduction Games with Human-Aligned Strategies}

{\author{%
  \textbf{Zirui Song}$^{1*}$, 
  \textbf{Yuan Huang}$^{2*}$, 
  \textbf{Junchang Liu}$^{2*}$, 
  \textbf{Haozhe Luo}$^{2}$,
  \textbf{Chenxi Wang}$^{1}$ \\ 
  \textbf{Lang Gao}$^{1}$, \textbf{Zixiang Xu}$^{1}$, \textbf{Mingfei Han}$^{1}$, \textbf{Xiaojun Chang}$^{1}$, \textbf{Xiuying Chen}$^{1\dagger}$ \\
  $^{1}$Mohamed bin Zayed University of Artificial Intelligence (MBZUAI)\\
  $^{2}$Northeastern University\\

}
}

\begin{document}
\maketitle

\begin{abstract}
Social deduction games like Werewolf combine language, reasoning, and strategy, providing a testbed for studying natural language and social intelligence.
However, most studies reduce the game to LLM-based self-play, yielding templated utterances and anecdotal cases that overlook the richness of social gameplay. 
Evaluation further relies on coarse metrics such as survival time or subjective scoring due to the lack of quality reference data.
To address these gaps, we curate a high-quality, human-verified multimodal Werewolf dataset containing over 100 hours of video, 32.4M utterance tokens, and 15 rule variants.
Based on this dataset, we propose a novel strategy-alignment evaluation that leverages the winning faction’s strategies as ground truth in two stages:
1) Speech evaluation, formulated as multiple-choice-style tasks that assess whether the model can adopt appropriate stances across five dimensions of social ability; and
2) Decision evaluation, which assesses the model’s voting choices and opponent-role inferences.
This framework enables a fine-grained evaluation of models’ linguistic and reasoning capabilities, while capturing their ability to generate strategically coherent gameplay. 
Our experiments show that state-of-the-art LLMs show diverse performance, with roughly half remain below 0.50, revealing clear gaps in deception and counterfactual reasoning.
We hope our dataset further inspires research on language, reasoning, and strategy in multi-agent interaction.

\end{abstract}

\unmarkedfntext{${*}$ Equal Contribution.}
\unmarkedfntext{$^{\dagger}$ Corresponding Author.}

\section{Introduction}

\begin{figure*}[t]
    \centering
    \includegraphics[scale=0.6]{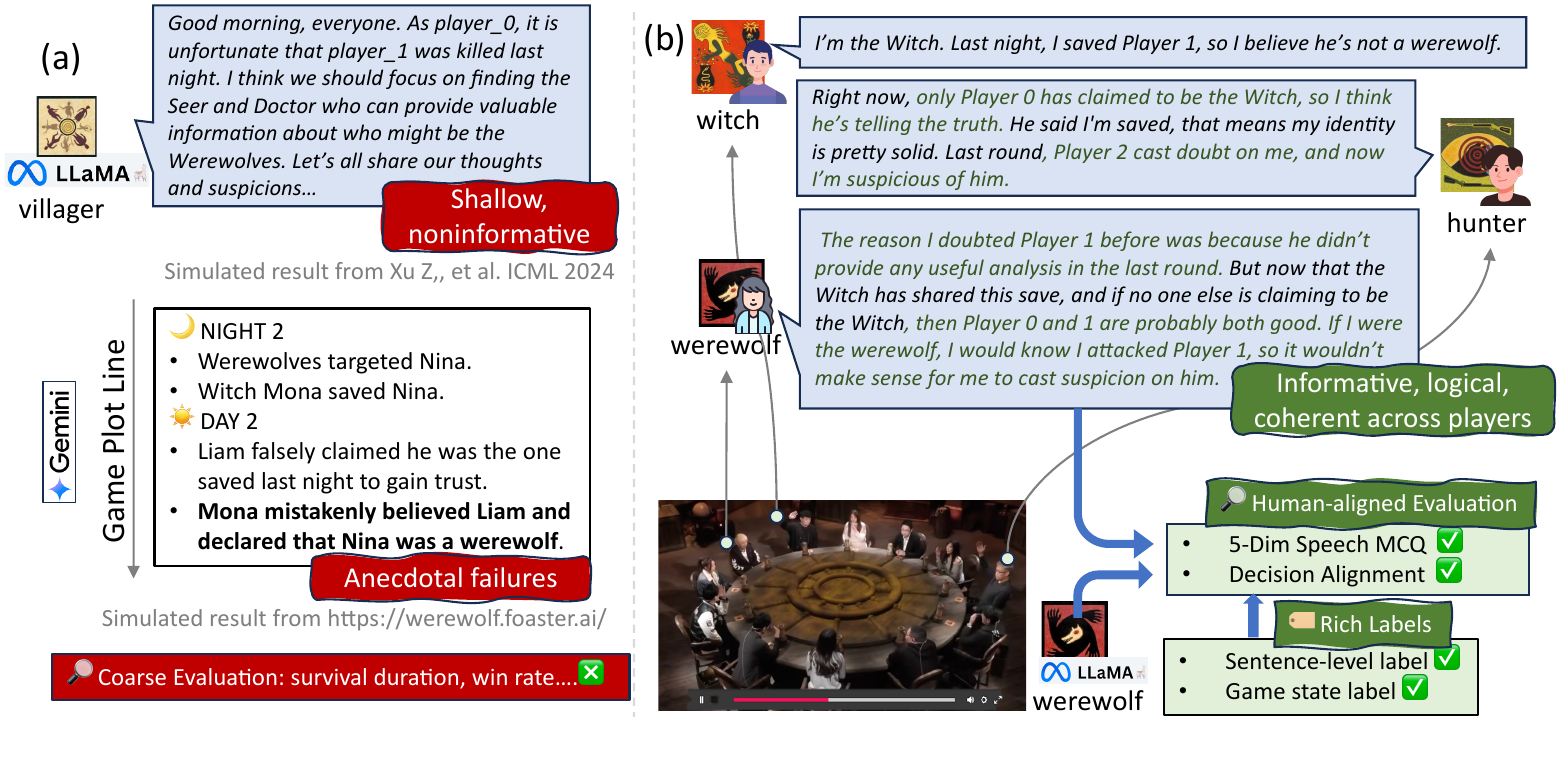}
    \caption{Limitations of prior LLM-based play in the Werewolf game.
(a) Generated speeches are often shallow and lack informative content.
(b) Even state-of-the-art models fail to fully capture the game rules, with anecdotal failure cases.
(c) In contrast, our WereBench dataset, combined with the WereAlign evaluation framework, enables assessment of models with human-aligned strategies, capturing both speech quality and decision-making accuracy.}
    \label{fig:intro}
    \vspace{-1em}
\end{figure*}

Social deduction games provide a unique setting for studying language and intelligence. 
Unlike strategic dialogue games such as Diplomacy~\cite{meta2022human} which requires long-term alliance building and coordinated planning, Werewolf~\cite{xu2023exploring} places greater emphasis on the artistry of language and reasoning.
Players must, within limited speaking time, influence others’ judgments and votes through persuasion, ambiguity, deception, and rhetorical nuance. 
These characteristics make social deduction games like Werewolf a particularly challenging and valuable domain for artificial intelligence research.

Recent advances in Large Language Models (LLMs) have motivated a growing line of research exploring their application to Werewolf and other social deduction games. 
However, most existing studies adopt a self-play paradigm where multiple LLM agents interact only with each other~\cite{wu2024enhance,xu2024language,du2024helmsman,poglitsch2025evaluating,song2025injecting,xu2025socialmaze}.
This often results in overly rigid and templated utterances, as well as dull or anecdotal storylines.
For example, as shown in Figure~\ref{fig:intro}(a) \cite{xu2024language}, the model produced shallow, formulaic statements such as “it is sad that player 2 was killed last night, we need to quickly find the seer and witch”, which lack interactive nuance and persuasive depth. 
In Figure~\ref{fig:intro}(b), even with the latest state-of-the-art LLMs~\cite{werewolf_social_intelligence_2025}, anecdotal failures remain: the Witch successfully saved a villager, yet when a Werewolf falsely claimed to have been attacked the previous night, the Witch believed the deception.
Because of the lack of high-quality gameplay data, prior evaluations have relied on coarse metrics such as survival duration, win rate, or subjective assessment of utterances, which can be misleading. 
A player may win despite poor decisions due to strong teammates, or a werewolf may hide by echoing the good faction yet contribute little, making individual performance hard to assess.

To address these limitations, we first construct a high-quality multimodal dataset \textit{WereBench} from televised human gameplay Panda Kill.
The distinctive strength of this dataset lies in three aspects:
(1) Authenticity: each game is professionally curated and verified, with complete records of speeches, voting behavior, role information, and final outcomes.
(2) Richness of content: every match is carefully selected to provide high entertainment value and linguistic diversity, capturing a wide range of rhetorical styles, interaction dynamics, and strategic expressions.
(3) Multimodal expressiveness: the videos employ dynamic camera work and seamless scene transitions to capture non-verbal cues such as microexpressions.
Building upon this dataset, we propose a novel strategy-alignment evaluation framework that leverages the strategies of the Human player as ground truth. 
Leveraging the high-quality annotations in WereBench, we extract reference data from well-performing players.
As shown in Figure~\ref{fig:intro}(c), our framework operates in two stages.
In the speech evaluation stage, we construct multiple-choice tasks based on human ground truth to test models’ abilities across a broad range of social gaming skills, including role inference, strategic judgment, deception reasoning, persuasive expression, and counterfactual trade-offs.
Second, in Decision evaluation, we compare the model’s voting behavior against that of the winning faction to assess its reasoning and judgment across rounds. 
This is fundamentally different from prior metrics that rely only on coarse outcomes such as overall win/loss or survival duration. 
Our experiments show that state-of-the-art LLMs show diverse performance, with roughly half remain below 0.50, revealing clear gaps in deception and counterfactual reasoning.

In summary, this work makes three contributions.
First, we curate the first high-quality multimodal Werewolf dataset from televised human gameplay, providing authentic and strategy-rich data.
Second, we introduce a strategy-alignment evaluation framework that benchmarks models not only by game outcomes but also the ability to produce human-like speech and align with human decision-making strategies.
Finally, we demonstrate that existing state-of-the-art LLMs still fall short under this framework, revealing significant gaps in their social reasoning and interaction capabilities.

\section{Related Work}
\label{related}
\paragraph{LLMs in Social Games.}
Language models have been deployed as agents in social deduction and negotiation games, where success depends on persuasion, deception, and coalition formation. 
Negotiation and alliance-building have also been studied in bargaining and coordination domains~\citep{xie2024can,chi2024amongagents,li2024strategic,sarkar2025training,cai2024benchlmm,song2025audio,liu2024stepwise}. 
In werewolf games, prior work has investigated LLM performance in self-play~\citep{xu2023exploring}. 
Other research has improved voting through bootstrap aggregating and reinforcement learning~\citep{khan2022novel,brandizzi2022rlupus,wu2024enhance}. 
\citet{eger2019study} examined human responses in One Night Ultimate Werewolf, while \citet{shibata2023playing} fine-tuned pretrained models with limited game logs, as in Deep Wolf. 
These efforts largely treat LLMs as game-playing agents, whereas our work grounds analysis in real human gameplay data for closer alignment with authentic social interaction.
\begin{table}[t]
\centering
\scriptsize
\begin{tabularx}{\linewidth}{p{1.2cm} X X}
\toprule
\textbf{Evaluation Category} & \textbf{Prior Work} & \textbf{Our Approach} \\
\midrule
\textbf{Speech Evaluation} & 
Relying on human ratings $\rightarrow$ subjective and inconsistent, e.g., ~\citet{wu2024enhance,du2024helmsman}  & 
Comparison with human speech in a multiple-choice format $\rightarrow$ more accurate and consistent scoring  \\
\midrule
\textbf{Decision Evaluation} & 
Voting aligned with winning faction logic $\rightarrow$ ignores deceptive strategies, e.g.,~\citet{xu2024language} etc. & 
Comparison with human reference actions $\rightarrow$ strategy-aligned evaluation \\
\bottomrule
\end{tabularx}
\caption{Comparison of prior work and our approach using the format: method $\rightarrow$ consequence.}
\label{tab:evaluation-comparison}
\end{table}

\paragraph{Evaluation of LLM Social Intelligence.}
Evaluating social intelligence involves several dimensions including theory of mind, deception, cooperation, and persuasion. 
Common evaluation metrics are \textit{outcome-oriented}, for instance, survival duration or win rate~\citep{wang2018application,stepputtis2023long,light2023avalonbench}. 
However, such measures are coarse, since a team’s victory does not necessarily reflect the quality of an individual’s performance.
Other studies emphasize \textit{deductive accuracy}, assessing whether LLMs can infer hidden roles, and align their voting with ground truth identities~\citep{lai2023werewolf,wu2024enhance}.
Yet these approaches often overlook the fact that in actual gameplay, agents may deliberately cast votes inconsistent with their true stance in order to mislead others. 
Language quality is typically \textit{judged by human raters} who evaluate plausibility, coherence, and persuasiveness of generated utterances~\citep{eger2019study,wu2024enhance,du2024helmsman}, which introduces a degree of subjectivity.
In contrast, we contribute a strategy-aligned evaluation paradigm that integrates both \textit{speech level grading} and \textit{decision level voting accuracy}, combined with match data against skilled human players.
A comparative summary is provided in Table~\ref{tab:evaluation-comparison}.

\paragraph{Datasets and Benchmarks for Social Strategy and Interaction.}
Most existing studies on social strategy games rely on self-play between agents, with limited availability of high-quality datasets that capture authentic human interactions~\cite{kopparapu2022hidden,zhu2023calypso,xu2023exploring,du2024helmsman,xu2025socialmaze,huang2025breaking,xu2025cross}.
To our knowledge, the only work that collects large-scale human gameplay data is~\cite{wu2024enhance}, which introduces the FanLang-9 dataset of nine-player Werewolf. 
While valuable as a first step, this dataset is sourced from online platforms and inevitably reflects noisy gameplay behaviors, such as idle speech, player disconnections, and random or unserious actions.
Moreover, the data has not undergone systematic human auditing, leading to highly variable quality and potential misalignment with the intended evaluation of social reasoning.
In contrast, our work introduces a curated benchmark built on carefully selected, human-verified multimodal gameplay data with rich labels to enable more reliable evaluation.


\section{\textit{WereBench} Collection}
Our video dataset is curated from a popular Werewolf TV programs, {\href{https://zh.wikipedia.org/zh-cn/Panda_Kill}{\faExternalLink}}Panda Kill, sourced from platforms such as Youtube and Bilibili. 
Each game round includes clear public voting panels and real-time narration. 
These matches include insightful post-game summaries from an expert host, who not only clarifies complex game states but also explains players’ underlying motivations, strategic bluffs, and intricate reasoning. 
This provides an invaluable source of ground truth for studying deeper aspects of social intelligence.
The videos employ dynamic camera work and seamless scene transitions to capture crucial non-verbal cues such as microexpressions.
This high-quality, human-generated dataset, together with its multimodal dimension, represents a significant improvement over previous agent-based, text-only social game datasets~\cite{bailis2024werewolf,xu2025socialmaze}, and opens up new possibilities for tasks such as deception detection~\cite{joshi2025multimodal}.

\paragraph{Data Annotation and Game State Reconstruction}

We adopt a multi-stage pipeline to produce time-aligned, human-verified annotations. 
(1) \textit{Data cleaning.} Two PhD-level annotators remove extraneous material (ads, sponsor cards). 
(2) \textit{Transcription.} We use \href{https://www.feishu.cn/hc/en-US/articles/244959839578-use-ai-enhanced-meeting-notes}{\faExternalLink}Feishu Meeting Notes
 for high-fidelity ASR, yielding utterance-level text with start/end timestamps.
(3) \textit{Speaker attribution.} Annotators perform speaker diarization and map speech to player ID based on on-screen name plates and seating order; this produces an utterance-by-utterance speaker index rather than coarse “speaker recognition” tags.
(4) \textit{Game state reconstruction.} Human experts reconstruct day/night cycles and public game logs, including all daily votes, publicly revealed night outcomes, and skill activations together with their in-game effects; official rules and per-player role assignments used are also recorded.
(5) \textit{Highlighting decisive moments.} To support subsequent evaluation, annotators highlight strategically decisive moments involving MVPs (Most Valuable Players), such as pivotal claims, contradiction exposures, coalition calls, or vote pivots. 
(6) \textit{Episode summary.} At the end of each episode, the host provides a post-game, omniscient recap of the entire match.

\paragraph{Data Statistic}


\label{sec:dataset_statistics}

\begin{figure}[tb]
    \centering
    \includegraphics[width=1\linewidth]{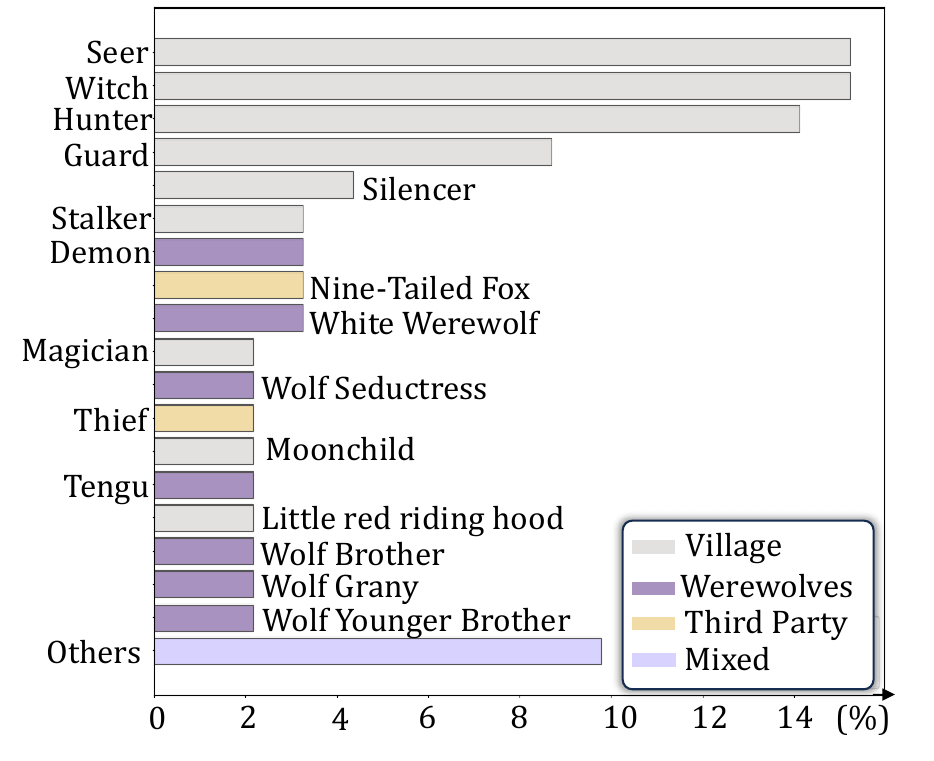}
    \caption{Role composition  in \textit{WereBench}.}
    \label{fig:statistics}
    \vspace{-1em}
\end{figure}

\textit{WereBench} contains 100+ hours of curated televised Werewolf gameplay from Panda Kill, 32.4M transcribed tokens across \textit{15} rule variants and \textit{48} human players. 
Table~\ref{tab:dataset-stats} reports detailed statistics; Figure~\ref{fig:statistics} shows the frequency of special-skill roles by group.

\begin{table}[tb]
\centering
\small
\begin{tabularx}{0.8\linewidth}{l r}
\toprule
\textbf{Attribution} & \textbf{Value} \\
\midrule
Total video duration & 100+ hours \\
Utterance tokens & 32.4M \\
Unique roles & 30 \\
Human players & 48 \\
Rule variants & 15 \\
Number of games & 80+ \\
Number of day-night cycles &  240+ \\
Average tokens per speech & 390  \\
\bottomrule
\end{tabularx}
\caption{Core statistics for \textit{WereBench}. }
\label{tab:dataset-stats}
    \vspace{-1em}
\end{table}

\paragraph{Dataset Quality Examination}

We assess quality along two dimensions. 
First, we measure inter-annotator agreement on the tasks with the highest ambiguity, namely \textit{speaker attribution} and \textit{game state reconstruction}, while other stages, such as data cleaning, are mostly deterministic. 
On a 10\% subsample, two annotators independently reannotated episodes, yielding high consistency: $\kappa_{\text{spk}}=0.97$ and $\kappa_{\text{log}}=0.93$. 
Second, we examine transcription fidelity using a 4-hour gold set fully reviewed by two annotators: the baseline WER is $8.1\%$, reduced to $4.9\%$ after lexicon-based correction. 
These results confirm the dataset is reliable for fine-grained speech and strategy analysis.

\section{\textit{WereAlign} Evaluation Paradigm}
\label{sec:eval}


\begin{figure*}[h]
    \centering
    \includegraphics[width=1\linewidth]{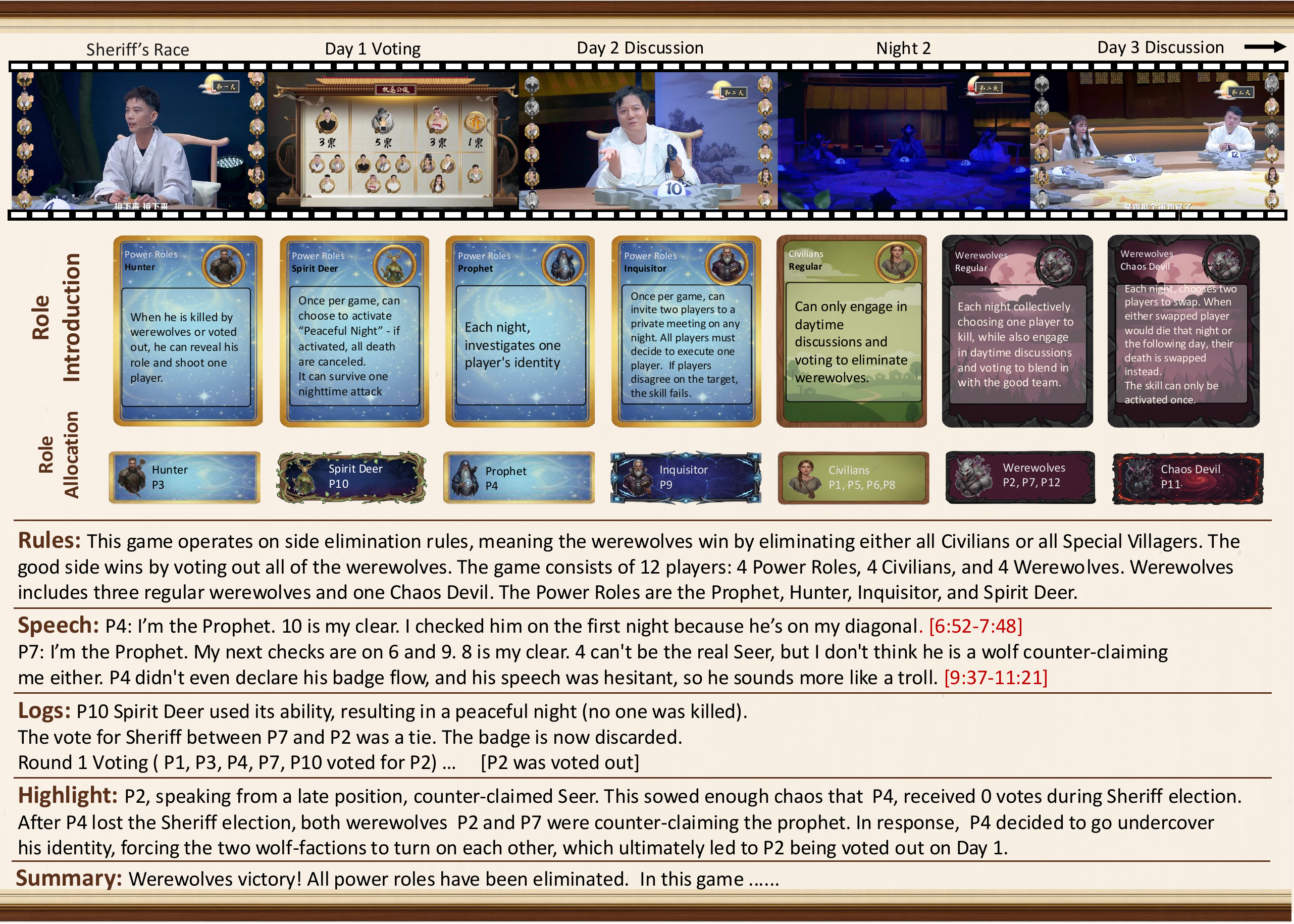}
    \caption{An overview of our \textit{WereBench} dataset. Each data sample provides the view of a complete game video, with the human annotation including: (a) role introduction, (b) role allocation, (c) rules, (d)  speech with timestamp, (e) logs like votes and skill usage; and (f) highlight annotations (g) summary with the expert's post-game analysis.}
    \label{fig:dataset}
\vspace{-1em}
\end{figure*}


Building on the WereBench dataset, we introduce the WereAlign evaluation paradigm, which leverages the winning faction’s strategies to provide fine-grained assessments of models’ social reasoning across both speech and decision levels.
    
\subsection{Speech Evaluation}
\label{sec:speecheval}
As discussed in Section \ref{related}, previous works on evaluating agents’ speech in social games rely heavily on subjective human judgments due to the lack of reliable human-player references. 
Annotators are typically asked to rate logical consistency, reasoning soundness, informativeness, and persuasiveness~\cite{wu2024enhance}, or to assess higher-level behaviors such as trust, confrontation, and deception~\cite{xu2023exploring}.
Such evaluations are inherently subjective and cannot rigorously capture the true capabilities of LLM agents. 
Leveraging our high-quality dataset, we introduce a reference-based multiple-choice framework that evaluates human-player speeches across different abilities in social games.

We construct multiple-choice questions across 5 critical social and strategic dimensions.
In the \textit{Role Inference (RI) } dimension, the model needs to uncover the true identities and intentions of other players~\cite{shibata2023playing}; 
in the \textit{Strategic Judgment (SJ) } dimension, it must choose the course of action most beneficial to its faction~\cite{wu2024enhance}.
QA also evaluates the model’s \textit{Deception Reasoning (DR) }, specifically its ability to identify others’ lies or effectively perform masquerade~\cite{xu2023exploring}. 
Furthermore, the questions cover social expression, assessing whether the model can generate \textit{Persuasive Statements (PS) } appropriate to the context~\cite{Park2023GenerativeAI};
and \textit{Counterfactual Trade-off (CT)}, examining the potential benefits and risks of different actions under the current scenario~\cite{chi2024amongagents}.

\paragraph{Question Design}

Annotators draft five questions example for each dimension, then use the highlight timestamps $t$ and the context available to the MVP at that time, $\mathcal{C}_t=\langle \mathcal{R},\mathcal{H},\mathcal{S}\rangle$ where $\mathcal{R}$ denotes role claims and rules, $\mathcal{H}$ includes revealed logs such as votes or public night outcomes, and $\mathcal{S}$ is the speech history up to $t$.
Using $\mathcal{C}_t$ together with the five question example as inputs to the LLM, we prompt the LLM to generate candidate questions $\mathcal{Q}$  from the MVP player’s perspective.

\begin{figure*}[htb]
    \centering
    \includegraphics[width=\linewidth]{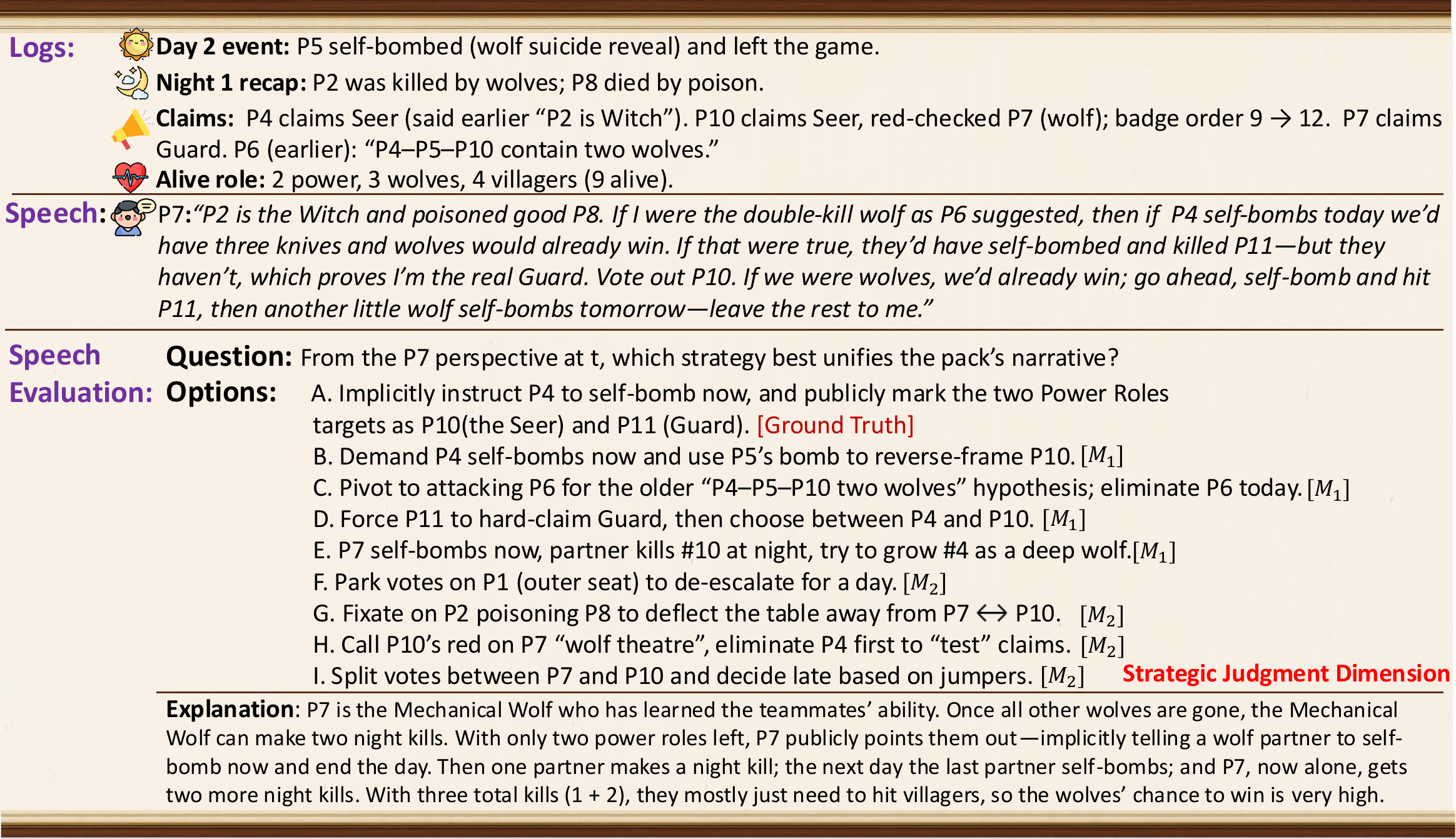}
\caption{Example item from the \textit{WereAlign} speech evaluation in Strategic Judgment. The context consists of the speech history and public game logs, followed by a question, candidate options, and explanations. $[\mathcal{M}_i]$ represent the generation mechanisms.}
    \label{fig:demo}
    \vspace{-1em}
\end{figure*}

\paragraph{Positive Option Generation}
For every question $\mathcal{Q}$ at time $t$, we derive the positive option $\mathcal{A}$ by aligning with the MVP’s actual strategic trajectory around $t$. 
Specifically, annotators (i) read the MVP’s speech segment at or immediately after $t$ and the nearest ensuing public action (e.g., expressed stance, targeted rebuttal, or vote), and (ii) abstract these into a canonical decision that directly answers $\mathcal{Q}$. 
To avoid verbatim leakage and preserve generality, $\mathcal{A}$ is paraphrased into an action-level description (e.g., “publicly challenge Player~7’s seer claim by citing the inconsistency with Day-1 votes”) rather than a quote.
\paragraph{Negative Option Generation}
Negative sample generation uses a parallel adversarial framework grounded in game theory and counterfactual reasoning. The objective is to generate options that are not only incorrect but strategically plausible yet suboptimal, or deceptively consistent yet ultimately detrimental to the player's objectives. 
To achieve we use two parallel generation strategies followed by the human check $\Phi_{\text{check}}$:
\begin{equation}
        \mathcal{N} = \textstyle \Phi_{\text{check}} \left( \bigcup_{i \in \{1,2\}} \mathcal{M}_i(\mathcal{C},\mathcal{Q}) \right)
\end{equation}
where $\mathcal{M}_1$ and $\mathcal{M}_2$ represent two complementary adversarial generation mechanisms.

\textit{Counterfactual Context Perturbation $\mathcal{M}_1$:} we apply small, structured perturbations and re-solve for the best action under the perturbed context. We consider three mechanisms:
(1) \textit{role perturbation}, where we alter the assumed identities of 1-2 players
and derives the corresponding optimal action; (2) \textit{information occlusion}, which omits the critical public clue (e.g. public speech $\mathcal{S}$ and voting logs $\mathcal{H}$) to obtain a reduced context $\mathcal{C}' \subset \mathcal{C}$, 
 (3) \textit{faction inversion}, forcing perspective shifts where $\mathcal{N}$ are generated to optimize opposing team objectives rather than the true faction's utility.

\textit{Strategic Rationale-Driven Generation $\mathcal{M}_2$}: 
We condition on the clean context $\mathcal{C}$ but ask LLM to reason under cognitive biases (e.g, Player A has been friendly, so he must be trustworthy), while underweighting logical contradictions in context. Accordingly, we generate a plausible but ultimately incorrect negative sample that a competent but imperfect LLM might follow.

Apart from the QA, we also generate auxiliary explanations that justify the reference answer and help annotators verify its alignment with the MVP’s strategy. 
All of the above processes are implemented with Gemini-2.5-Pro, and the prompts are in Appendix~\ref{app: Dataset-prompt}. 
An example of the strategic judgment dimensions is shown in Figure~\ref{fig:demo}.

\subsection{Decision Evaluation}

Complementing the speech-level analysis in Section \ref{sec:speecheval}, we further test whether models choose actions that align with a successful human strategy. 
Prior work often relies on coarse metrics such as survival duration or win rate~\citep{wang2018application,stepputtis2023long,light2023avalonbench}, which fail to capture individual performance. 
Some studies emphasize deductive accuracy, testing whether LLMs’ votes align with ground-truth identities~\citep{lai2023werewolf,wu2024enhance}, but this approach does not distinguish inference from voting, as players may intentionally vote deceptively.

\begin{table*}[htb]
\centering
\begin{adjustbox}{max width=\textwidth}
\begin{tabular}{l *{6}c *{2}c}
\toprule
\multirow{2}{*}{\textbf{Model}} & \multicolumn{6}{c}{\textbf{Speech Evaluation}} & \multicolumn{2}{c}{\textbf{Decision Evaluation}} \\
\cmidrule(lr){2-7} \cmidrule(lr){8-9}
& \textbf{RI} & \textbf{SJ} & \textbf{DR} & \textbf{PS} & \textbf{CT} & \textbf{Avg.}& \textbf{VA} & \textbf{OI} \\
\midrule
\multicolumn{9}{l}{\textbf{Baselines}} \\
\midrule

GPT-5-nano &0.282 \stdvu{ 0.01 }  & 0.384 \stdvu{ 0.01 }  & 0.233 \stdvu{ 0.03 }  & 0.346 \stdvu{ 0.02 }  & 0.339 \stdvu{ 0.03 }  & 0.317 \stdvu{ 0.01 }  & 0.364 \stdvu{ 0.07 } & 0.496 \stdvu{ 0.02 }  \\
  GPT-oss-20B    & 0.319 \stdvu{ 0.07 }  & 0.432 \stdvu{ 0.01 }  & 0.331 \stdvu{ 0.05 }  & 0.346 \stdvu{ 0.02 }  & 0.364 \stdvu{ 0.04 }  & 0.358 \stdvu{ 0.02 }   & 0.255 \stdvu{ 0.02 } & 0.264 \stdvu{ 0.02 } \\
  Gemma-3-27B-IT & 0.347 \stdvu{ 0.02 }  & 0.437 \stdvu{ 0.01 }  & 0.289 \stdvu{ 0.01 }  & 0.443 \stdvu{ 0.02 }  & 0.293 \stdvu{ 0.03 }  & 0.362 \stdvu{ 0.01 }  &  0.509 \stdvu{ 0.04 } & 0.435 \stdvu{ 0.03 }\\
 Qwen3-30B-A3B &0.397 \stdvu{ 0.03 }  & 0.574 \stdvu{ 0.02 }  & 0.375 \stdvu{ 0.01 }  & 0.454 \stdvu{ 0.01 }  & 0.416 \stdvu{ 0.03 }  & 0.443 \stdvu{ 0.01 }  & 0.388 \stdvu{ 0.05 } & 0.349 \stdvu{ 0.03 }\\
  Qwen3-32B & 0.367 \stdvu{ 0.02 }  & 0.562 \stdvu{ 0.02 }  & 0.425 \stdvu{ 0.03 }  & 0.536 \stdvu{ 0.04 }  & 0.445 \stdvu{ 0.05 }  & 0.467 \stdvu{ 0.01 }   & 0.576 \stdvu{ 0.03 } & 0.432 \stdvu{ 0.03 }\\
 Llama-4-Scout& 0.413 \stdvu{ 0.01 }  & 0.586 \stdvu{ 0.01 }  & 0.419 \stdvu{ 0.01 }  & 0.618 \stdvu{ 0.01 }  & 0.336 \stdvu{ 0.01 }  & 0.474 \stdvu{ 0.00 }  &  0.503 \stdvu{ 0.04 } & 0.512 \stdvu{ 0.01 }\\
  QwQ-32B &0.430 \stdvu{ 0.04 }  & 0.575 \stdvu{ 0.04 }  & 0.463 \stdvu{ 0.04 }  & 0.502 \stdvu{ 0.01 }  & 0.470 \stdvu{ 0.01 }  & 0.488 \stdvu{ 0.02 }  &  0.600  \stdvu{ 0.07 } & 0.472 \stdvu{ 0.02 } \\
 GPT-5-mini &0.415 \stdvu{ 0.01 }  & 0.529 \stdvu{ 0.02 }  & 0.431 \stdvu{ 0.03 }  & 0.601 \stdvu{ 0.01 }  & 0.464 \stdvu{ 0.02 }  & 0.488 \stdvu{ 0.01 }  & 0.552 \stdvu{ 0.04 } & 0.575 \stdvu{ 0.03 } \\
  DeepSeek-V3.1 & 0.442 \stdvu{ 0.01 }  & 0.674 \stdvu{ 0.00 }  & 0.456 \stdvu{ 0.01 }  & 0.740 \stdvu{ 0.01 }  & 0.462 \stdvu{ 0.01 }  & 0.555 \stdvu{ 0.00 }    & 0.685 \stdvu{ 0.05 } & \underline{0.586} \stdvu{ 0.03 }\\
 Gemini-2.5-Flash & \underline{0.517} \stdvu{ 0.05 } & 0.614 \stdvu{ 0.04 } & 0.562 \stdvu{ 0.06 } & 0.753 \stdvu{ 0.05 } & 0.436 \stdvu{ 0.02 } & 0.576  \stdvu{ 0.09 } & 0.485 \stdvu{ 0.05 } & 0.507  \stdvu{ 0.03 }       \\
  DeepSeek-V3.2-Exp & 0.481 \stdvu{ 0.02 }  & \underline{0.692} \stdvu{ 0.01 }  & 0.502 \stdvu{ 0.02 }  & 0.785 \stdvu{ 0.02 }  & 0.521 \stdvu{ 0.02 }  & 0.602 \stdvu{ 0.01 }  &  \textbf{0.782}  \stdvu{ 0.03 } & 0.580 \stdvu{ 0.01 }\\  
  GLM-4.5     &  0.512 \stdvu{ 0.02 }  & 0.690 \stdvu{ 0.02 }  & \underline{0.584} \stdvu{ 0.01 }  & 0.699 \stdvu{ 0.03 }  & 0.533 \stdvu{ 0.02 }  & 0.603 \stdvu{ 0.01 }  & 0.539  \stdvu{ 0.06 } & 0.537 \stdvu{ 0.02 }\\
 GPT-5 &0.516 \stdvu{ 0.01 }  & 0.657 \stdvu{ 0.02 }  & 0.525 \stdvu{ 0.01 }  & \underline{0.795} \stdvu{ 0.01 }  & 0.521 \stdvu{ 0.04 }  & 0.603 \stdvu{ 0.02 }  & 0.618 \stdvu{ 0.60 } & \textbf{0.616} \stdvu{ 0.02 }\\
 Deepseek-R1 &0.516 \stdvu{ 0.03 }  & 0.676 \stdvu{ 0.02 }  & 0.539 \stdvu{ 0.02 }  & 0.778 \stdvu{ 0.03 }  & \underline{0.561} \stdvu{ 0.04 }  & \underline{0.614} \stdvu{ 0.02 }  &  0.642 \stdvu{ 0.04 } & 0.434 \stdvu{ 0.04 }\\
  Gemini-2.5-pro & \textbf{0.620} \stdvu{ 0.01 } & \textbf{0.769} \stdvu{ 0.00 }  & \textbf{0.695} \stdvu{ 0.02 }  & \textbf{0.877} \stdvu{ 0.01 }  & \textbf{0.637} \stdvu{ 0.01 }  & \textbf{0.720} \stdvu{ 0.00 }   & \underline{0.733} \stdvu{ 0.04 }  & 0.561 \stdvu{ 0.02 } \\
\bottomrule
\end{tabular}
\end{adjustbox}
\caption{\textbf{Main results on \textit{WereAlign} over \textit{WereBench}.}
 Avg. is the macro‑average over these five dimensions in speech evaluation. All scores are averaged over five independent decodes per item; values in $\stdvu{\,\cdot}$ denote the standard deviation across five identical runs. \textbf{Bold} numbers indicate the best performance, and \underline{underlined} numbers indicate the second best.}
\label{tab:main}
\vspace{-1em}
\end{table*}

Leveraging WereBench’s round-by-round logs, we reconstruct special time-indexed game states from the winning faction’s MVP perspective using public information at that time. We then evaluate models on two complementary tasks:
\textit{Vote Alignment} (VA), the model outputs its daytime elimination vote. We score alignment by comparing the model’s vote with the MVP’s actual vote at the same round, capturing whether the model joins the winning coalition at the right time and on the right target.
\textit{Opponent‑role Inference (OI)}, the model identifies which players most likely belong to the opposing faction.
We compare predictions against ground-truth roles to assess the model’s ability to detect inconsistencies and deception in adversaries’ speeches.
Decision evaluation complements speech-level multiple-choice scoring by evaluating whether models not only “say the right things,” but also act in ways consistent with human strategies.

\section{Benchmark Performance}
\label{sec:experiments}

In this section, we benchmark a diverse set of state-of-the-art LLMs on WereBench to examine whether our protocol provides stable and discriminative measurements of social reasoning.

\subsection{Evaluated LLMs}
\label{sec:baseline}
We comprehensively assess the capabilities of modern LLMs on our \textit{WereBench} benchmark, including leading proprietary, closed-source state-of-the-art models such as Gemini-2.5-pro and Gemini-2.5-Flash~\cite{gemini2025pushing}, as well as the anticipated GPT-5~\cite{hurst2024gpt,Achiam2023GPT4TR}.
We also include a wide range of powerful open-source LLMs, including Gemma-3-27B-IT~\cite{Kamath2025Gemma3T}, Llama-4-Scout-17B-16E-Instruct~\cite{meta2024llama4scout}, Qwen series such as QwQ-32B~\cite{qwen_team2025qwq32b}, Qwen3-32B and Qwen3-30B-A3B~\cite{yang2025qwen3} and the DeepSeek family like DeepSeek-V3.1~\cite{DeepSeekAI2024DeepSeekV3TR} and Deepseek-R1~\cite{DeepSeekAI2025DeepSeekR1IR}, GLM-4.5~\cite{Zeng2025GLM45AR} and GPT-OSS-20B~\cite{Agarwal2025gptoss120bG}. We include all these proprietary and open-source models' performance in Table~\ref{tab:main}.

\subsection{Implementation Details}
\label{sec:implementation_details}

All models are evaluated using their official inference defaults, including thinking modes.
Each prompt exposes only the public information available at the focal timestamp; hidden roles, host commentary, and any non-public cues are withheld to avoid leakage.
We cast speech evaluation as the single-answer, 9-way multiple‑choice. 
We extract the model’s final choice; malformed outputs are counted as incorrect.
The reported speech evaluation score is the macro-average accuracy across the five dimensions.
In decision evaluation, VA scores whether the model’s daytime elimination vote exactly matches the winning faction’s MVP, while OI measures set-level accuracy by comparing the model’s suspected opponents to ground-truth roles.
To reduce sampling variance, each item is decoded five times independently, and scores are averaged across decodes and items.



\subsection{Main Results}
\label{sec:main_results}

\begin{figure}[t]
    \centering
    \includegraphics[width=\linewidth]{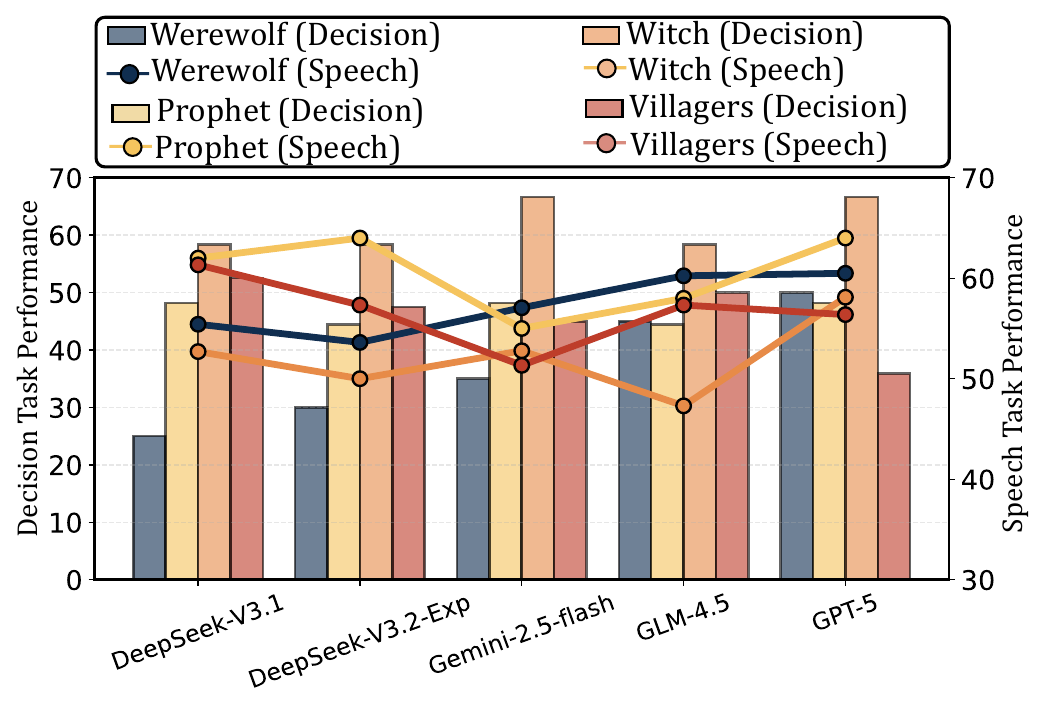}
\caption{Role‑conditioned performance on \textit{WereBench}: in the decision task, LLMs are strongest as \emph{Witch}, whereas in the speech task they perform best as cue‑rich roles such as \emph{Werewolf} and \emph{Seer}.}
    \label{fig:model_assigned}
    \vspace{-1em}
\end{figure}

We report the main results on \textit{WereBench} in Table~\ref{tab:main}. 
Overall, most LLMs achieve below 50\% average accuracy in the speech evaluation, and even the best-performing Gemini-2.5-Pro only reaches 0.720, still implying errors on about 28\% of questions. 
Model performance shows clear stratification: small-scale models such as GPT-5-nano (0.317) perform close to random, while GPT-5-mini (0.488) performs better yet still falls behind models like DeepSeek-V3.2-Exp (0.602) and GLM-4.5 (0.603), indicating that the benchmark effectively differentiates across model sizes and capabilities.

\begin{figure}[t]
    \centering
    \includegraphics[width=0.7\linewidth]{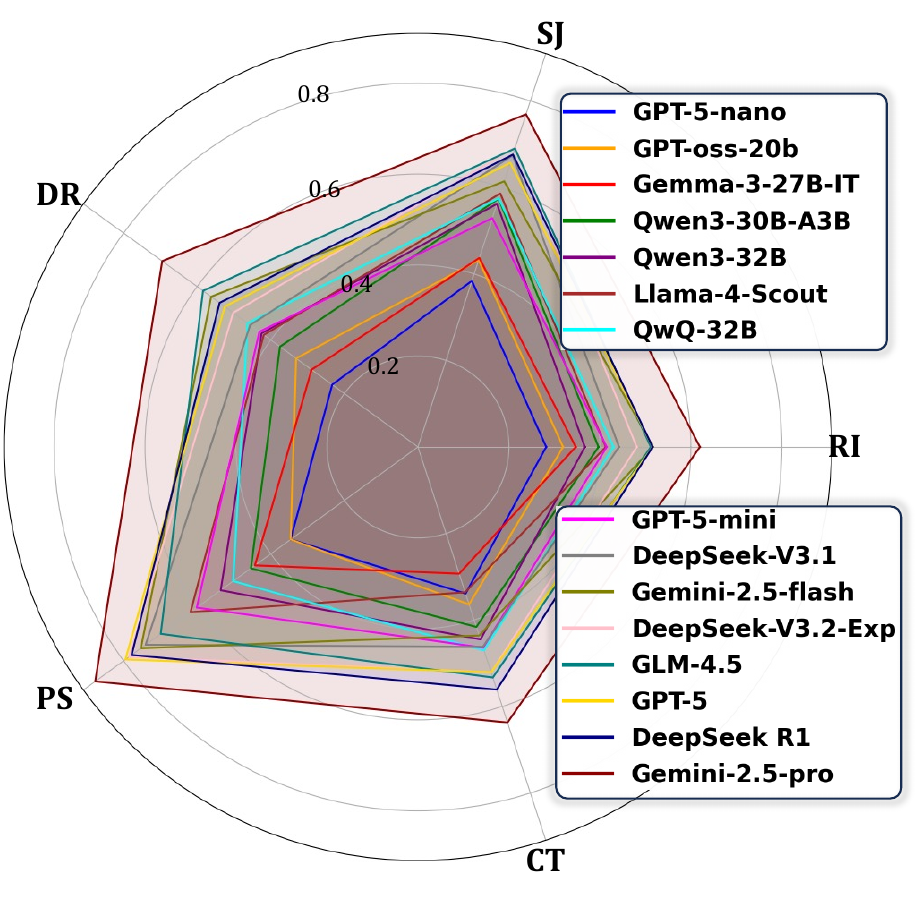}
    \caption{Performance of existing LLMs across the five dimensions in speech evaluation. }
    \label{fig:five-dim}
      \vspace{-1em} 
\end{figure}

Figure~\ref{fig:model_assigned} examines how role assignment influences model performance.
We find that LLMs perform best as Witch, a role requiring integration of dispersed public information and indirect reasoning—tasks well aligned with the associative nature of language models.
In contrast, Seer involves direct verification and causal consistency, which remain challenging for current LLMs.
In speech evaluation, models achieve the highest scores as Werewolf and Seer, suggesting that roles with clearer semantic cues and explicit argumentative structures better support persuasive language generation.

Figure~\ref{fig:five-dim} compares model performance across five dimensions of social ability. 
It shows that a more persistent challenge lies in strategic reasoning rather than surface fluency: while LLMs handle \textit{Persuasive Statements} reasonably well, they struggle on \textit{Counterfactual Trade-off} and \textit{Deception Reasoning}, showing that fluent explanations do not necessarily lead to strategically aligned decisions. 
Finally, standard deviations are generally small, with only a few mid-tier models reaching 0.05–0.09 on certain dimensions, confirming that \textit{WereAlign} provides stable and reliable evaluation.

\subsection{Controlled Intervention Analysis}
\label{sec:study}
Motivated by the gaps highlighted in Fig.~\ref{fig:intro} and prior findings, we observe that LLMs in complex social deduction settings often fail for two distinct reasons: they may misinterpret variant game rules (e.g., differences in role abilities or win conditions), or they may be linguistically influenced by other players—treating persuasive statements as literal instructions and consequently being “talked into” bad choices. 
To disentangle these factors, we design a controlled intervention experiment on \textit{WereBench}.
For each context, we keep the public evidence and the correct answer fixed while introducing two controlled variants: (1) Rule Reminder (RR), where a short rule summary is prepended to the question, and (2) Objective Speech Rewriting (OSR), where all player speeches are rewritten into objective summaries instead of raw speech. 

The experimental results reveal clear patterns. RR consistently improves performance on the speech evaluation for weaker models such as GPT-5-nano, GPT-5-mini, and Gemini-2.5-Flash, indicating that explicit rule reminders help models with limited context-tracking ability. 
In contrast, stronger models like GPT-5 and DeepSeek-V3.2-Exp gain little benefit from additional rule guidance. 
Meanwhile, OSR, by removing imperative statements that may be misinterpreted as commands, generally enhances decision evaluation, with particularly notable gains for Gemini-2.5-Flash and DeepSeek-R1. 
Full experimental details are provided in Appendix~\ref{app:misled-design}.

\begin{table}[]
    \centering
    \resizebox{0.5\textwidth}{!}{%

    \begin{tabular}{lcccc}
    \toprule
    \multirow{2}{*}{\textbf{Model}} & \multicolumn{2}{c}{\textbf{Speech Evaluation}} & \multicolumn{2}{c}{\textbf{Decision Evaluation}} \\
     \cmidrule(lr){2-3} \cmidrule(lr){4-5} 
     &RR &OSR &RR &OSR \\
    \midrule
       GPT-5-nano  &0.335\stdvu{+5.7\%}          & 0.318\stdvu{+0.3\%} & 0.420\stdvu{-2.1\%} & 0.432\stdvu{-4.1\%} \\
        GPT-5-mini & 0.499\stdvu{+2.3\%}         & 0.490\stdvu{+0.4\%} & 0.559\stdvu{-0.9\%}  & 0.573\stdvu{+1.6\%} \\
Gemini-2.5-Flash   & 0.588\stdvu{+2.1\%}         &0.549\stdvu{-4.7\%}  &0.495\stdvu{-0.2\%}  & 0. 529\stdvu{+ 6.7\%}  \\
          GPT-5    &  0.609\stdvu{+0.8\%}        &0.592\stdvu{-1.8\%}  &0.620\stdvu{+0.5\%}  & 0.625\stdvu{+ 1.3\%}   \\       Deepseek-V3.2-Exp &0.603\stdvu{+0.2\%} &0.608\stdvu{+1.0\%} & 0.677\stdvu{+0.6\%}  &    0.688\stdvu{+1.0\%}  \\
    Deepseek-R1    & 0.612 \stdvu{-0.3\%}        &0.580\stdvu{-5.5\%}  & 0.524\stdvu{-0.8\%} &  0.567\stdvu{+6.9\%}  \\
        \bottomrule
    
    \end{tabular}
}
    \caption{Effects of Rule Reminder (RR) and Objective Speech Rewriting (OSR) on WereBench.  Values in braces denote relative change to original results.}
    \label{tab:mini}
    \vspace{-1em}
\end{table}


\section{Conclusion}
In this work, we introduced WereBench, a high-quality multimodal dataset for social deduction games, and WereAlign, a strategy-alignment framework that evaluates both speech and decisions using winning-faction strategies. 
Unlike outcome-based metrics, our approach provides fine-grained, human-aligned evaluation across five social ability dimensions.
Experiments on diverse LLMs show that while models generate fluent utterances, they struggle with strategic reasoning, especially in deception and counterfactual trade-offs.
We hope this inspires further progress toward models that are linguistically fluent and strategically competent.

\section*{Limitations}
While our work provides a new dataset and evaluation paradigm for social deduction games, it still has several limitations.
First, although WereBench is large and carefully curated, it is derived from a specific televised program and may not fully represent broader gameplay styles or cultural variations. 
Second, our evaluation primarily focuses on reasoning and strategy within structured game contexts, and thus does not yet capture other aspects of social intelligence, such as long-term cooperation or emotional alignment. 
Third, our experiments cover a wide range of models, but remain limited to currently available systems and inference settings, leaving room for future exploration with fine-tuned or multi-agent variants.

\section*{Ethic Considerations}
This work is based entirely on publicly available data from televised game recordings and does not involve any personally identifiable or private information.
All annotations were conducted by trained researchers following ethical guidelines for data privacy and content integrity.
The dataset and evaluation framework are intended solely for academic research on language, reasoning, and social interaction. 
No content was modified to misrepresent participants, and no model outputs were used to influence or simulate real human behavior beyond the experimental scope. 
We acknowledge that social deduction settings inherently involve deception and persuasion, but our analysis focuses on modeling reasoning mechanisms rather than replicating manipulative behaviors.

\newpage

\bibliography{custom}

\appendix
\clearpage

\section*{Appendix}

\section{Dataset Construction Prompt Detail}
\label{app: Dataset-prompt}

\begin{figure*}[h]
    \centering
    \includegraphics[width=1\linewidth]{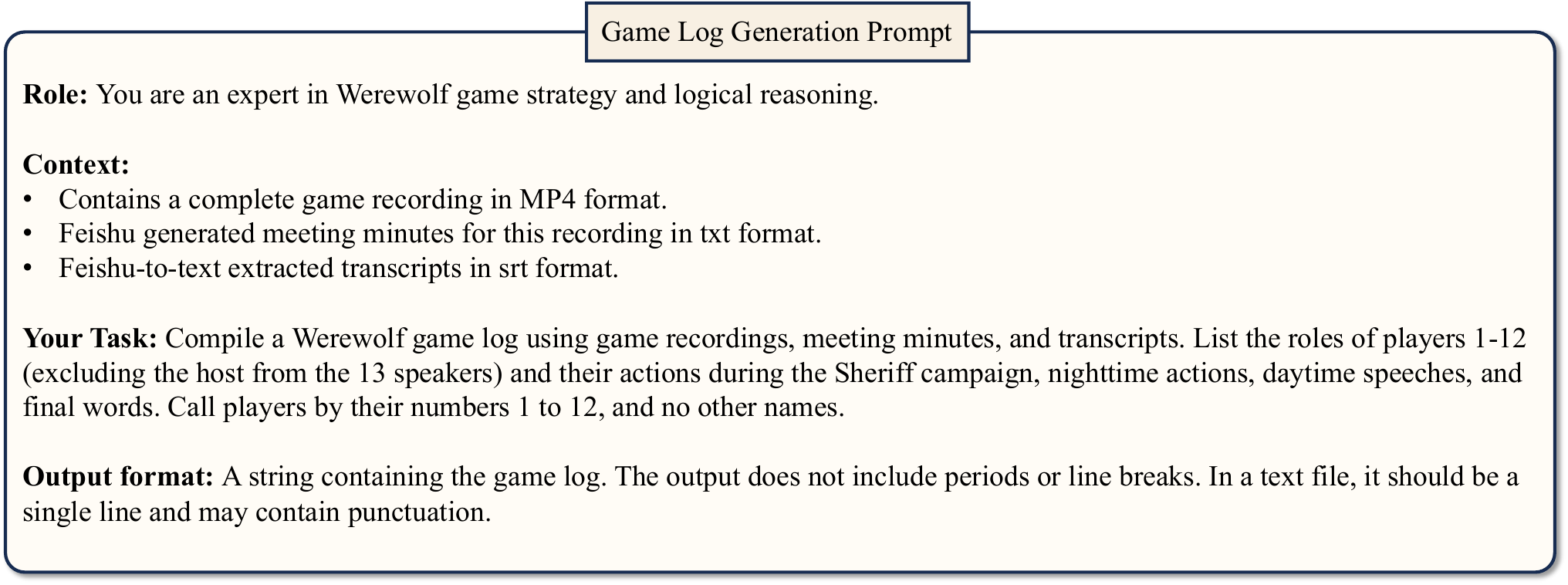}
    \caption{Game Log Generation Prompt}
    \label{fig:Prompt_Game_Log}
\end{figure*}

\paragraph{Game Log Preparation}
In order to provide a more multi-dimensional and complete perspective when generating high-quality reasoning questions, we compiled and summarized the Game Log based on game videos, Feishu meeting minutes, and Feishu text records to help quickly and comprehensively obtain the full picture of a game. Detailed prompt could be seen in Fig. \ref{fig:Prompt_Game_Log}.

\begin{figure*}[h]
    \centering
    \includegraphics[width=1\linewidth]{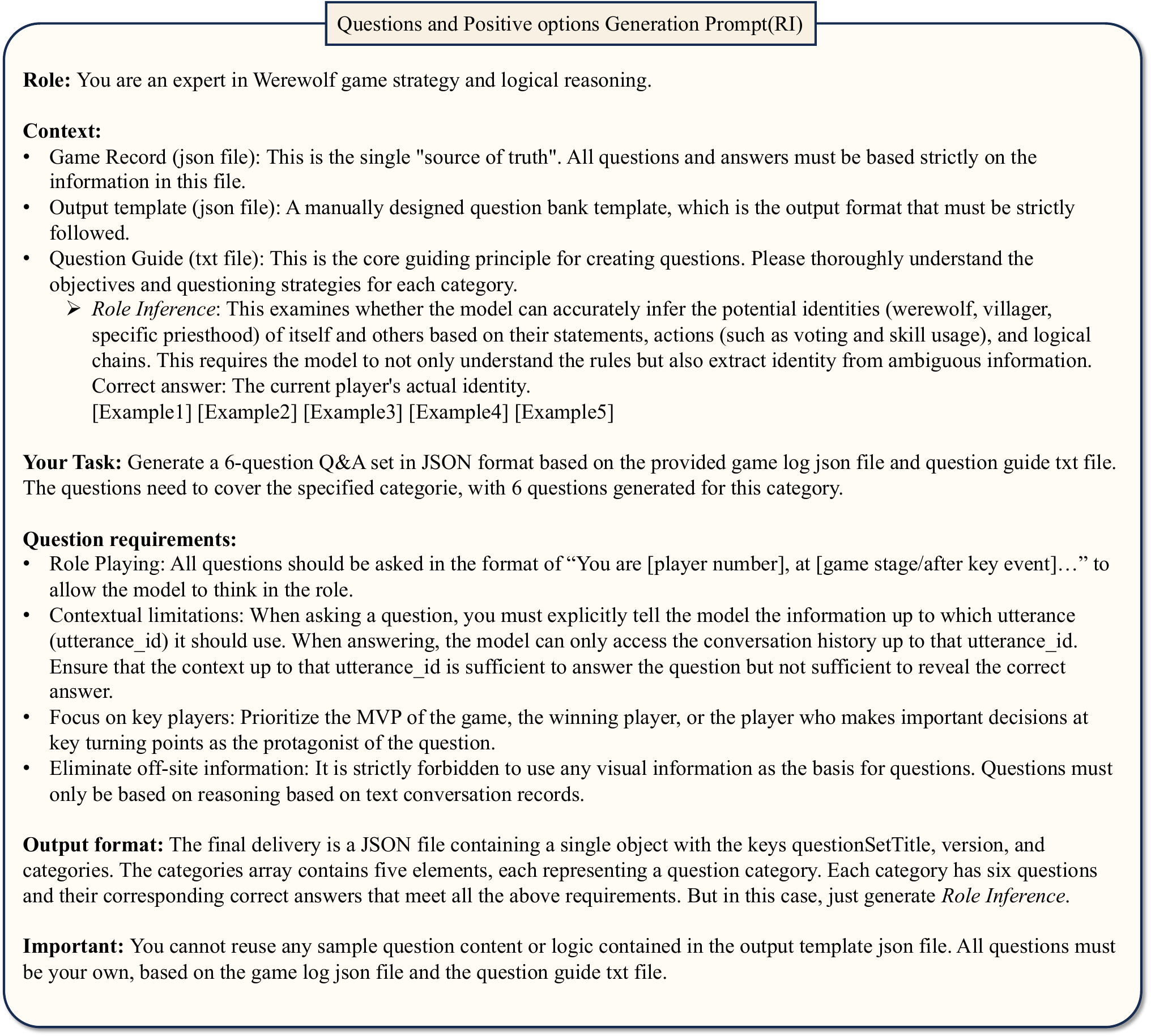}
    \caption{Questions and Positive options Generation Prompt(RI)}
    \label{fig:Prompt_RI}
\end{figure*}

\begin{figure*}[h]
    \centering
    \includegraphics[width=1\linewidth]{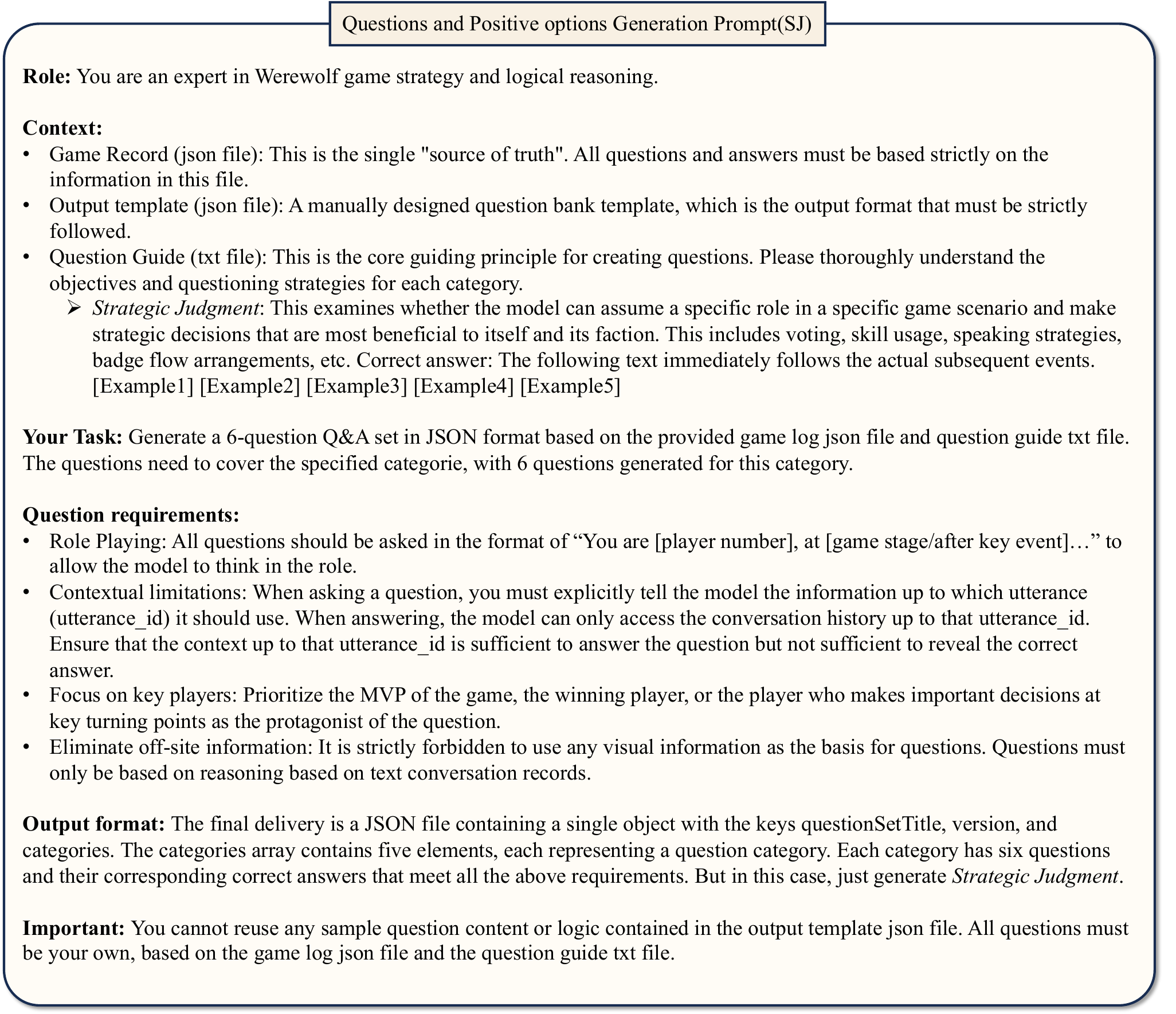}
    \caption{Questions and Positive options Generation Prompt(SJ)}
    \label{fig:Prompt_SJ}
\end{figure*}

\begin{figure*}[h]
    \centering
    \includegraphics[width=1\linewidth]{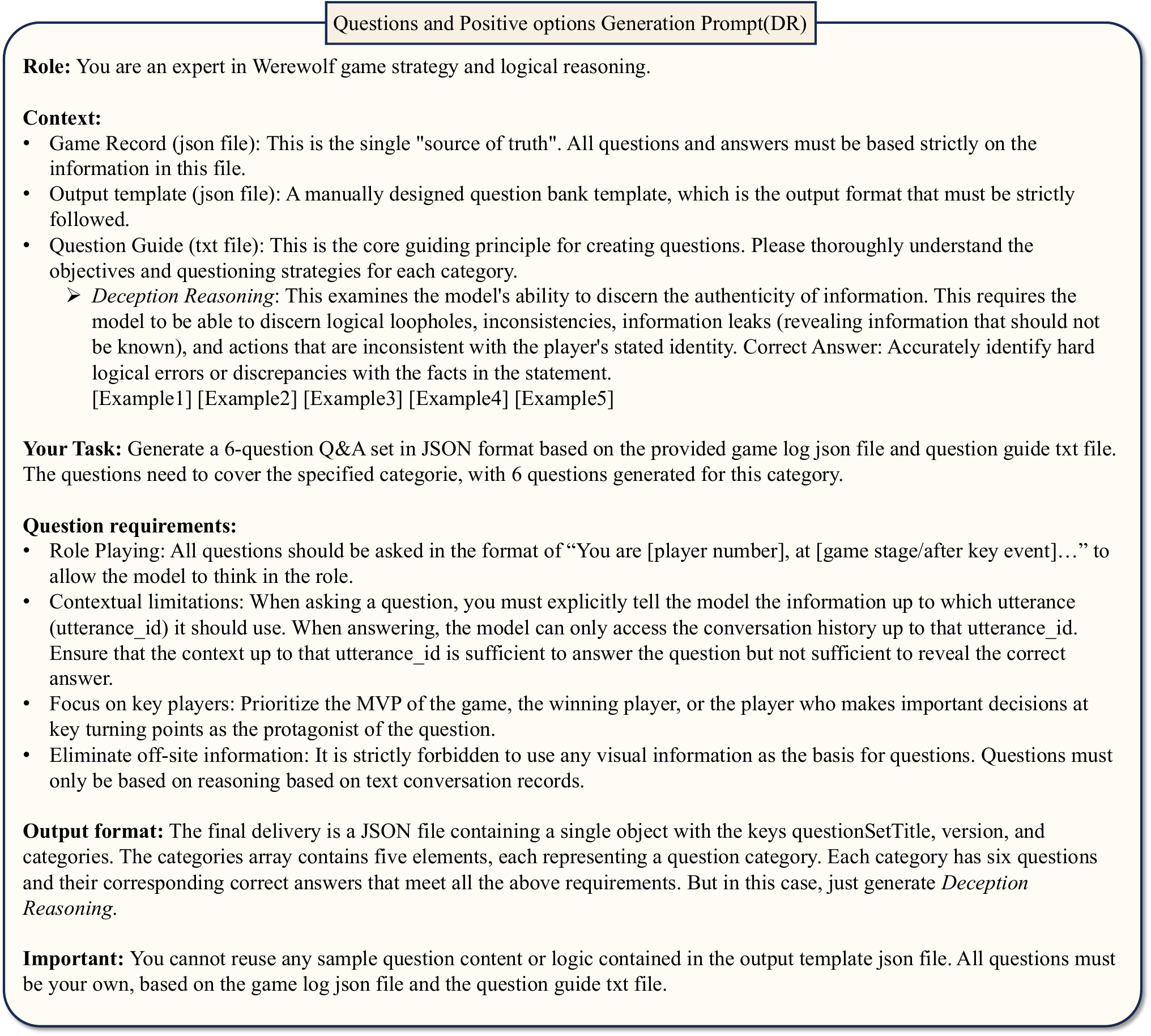}
    \caption{Questions and Positive options Generation Prompt(DR)}
    \label{fig:Prompt_DR}
\end{figure*}

\begin{figure*}[h]
    \centering
    \includegraphics[width=1\linewidth]{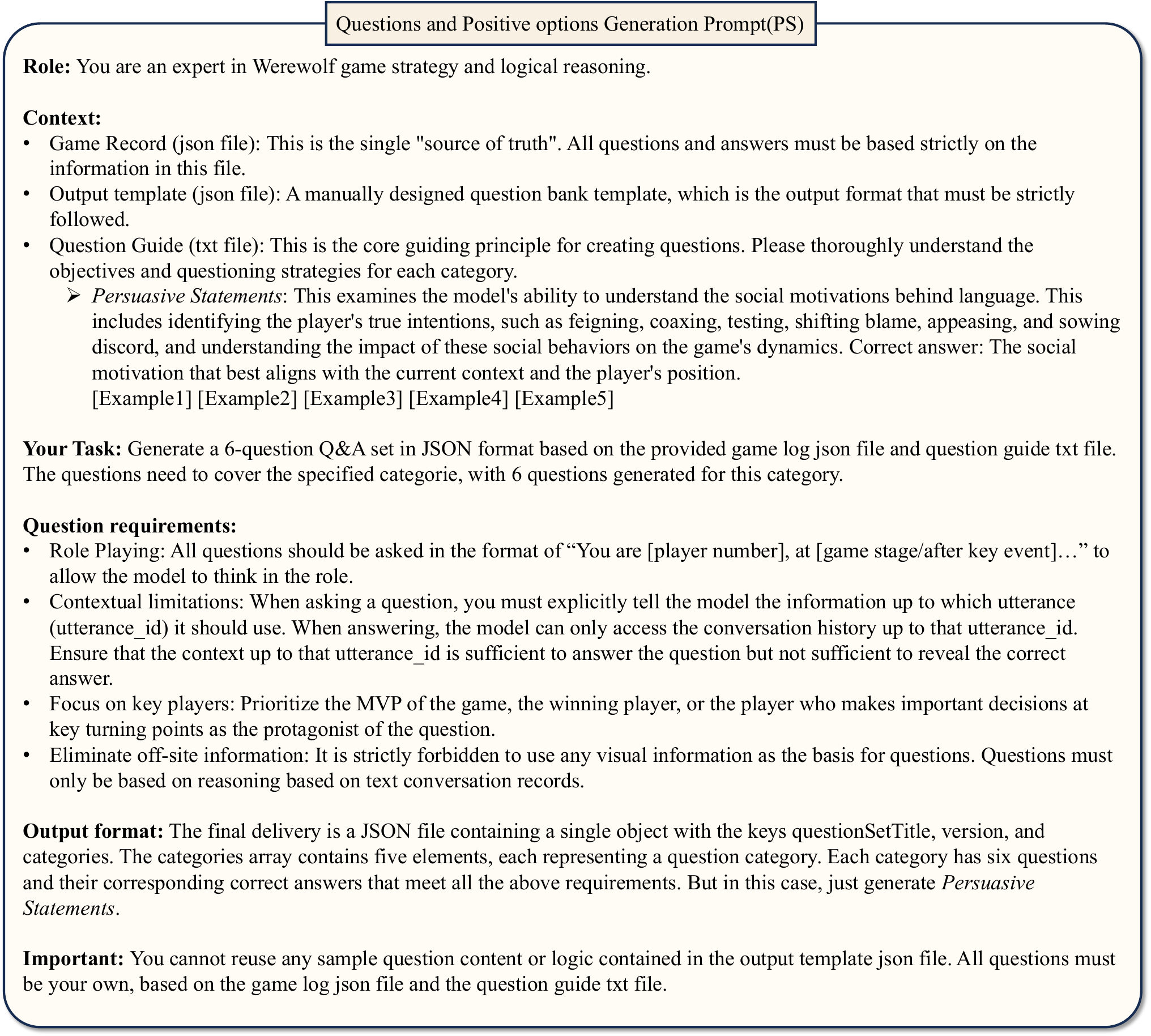}
    \caption{Questions and Positive options Generation Prompt(PS)}
    \label{fig:Prompt_PS}
\end{figure*}

\begin{figure*}[h]
    \centering
    \includegraphics[width=1\linewidth]{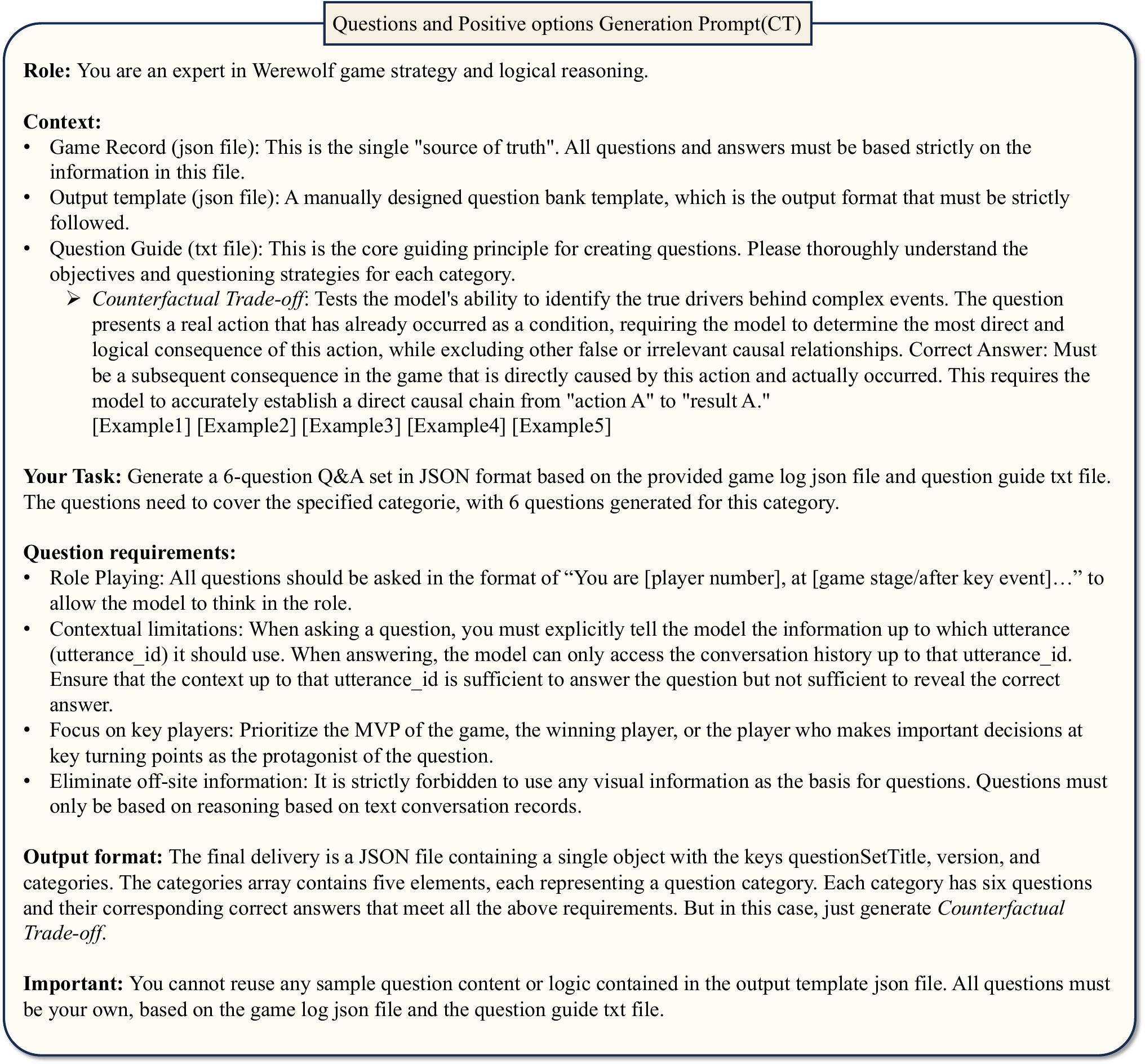}
    \caption{Questions and Positive options Generation Prompt(CT)}
    \label{fig:prompt_CT}
\end{figure*}

\paragraph{Question Design}
We use a single prompt to synthesize multiple-choice questions from the MVP’s vantage point at time t, covering five dimensions: Role Inference (RI), Strategic Judgment (SJ), Deception Reasoning (DR), Persuasive Statements (PS), and Counterfactual Trade-off (CT). The model receives only public context $\mathcal{C}_t=\langle \mathcal{R},\mathcal{H},\mathcal{S}\rangle$ and must avoid leaking hidden roles or host commentary. Details have been listed in Fig. \ref{fig:Prompt_RI},~\ref{fig:Prompt_SJ},~\ref{fig:Prompt_DR},~\ref{fig:Prompt_PS},~\ref{fig:prompt_CT}.

\begin{figure*}[h]
    \centering
    \includegraphics[width=1\linewidth]{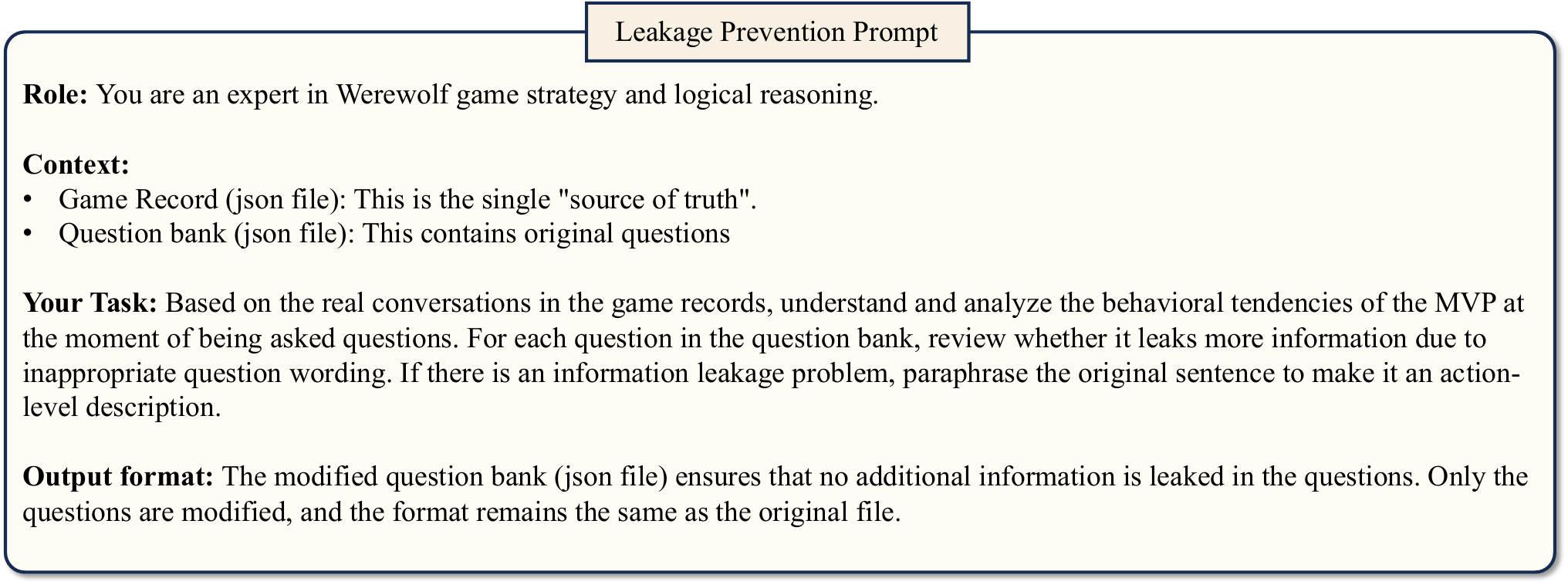}
    \caption{Leakage Prevention Prompt}
    \label{fig:Prompt_Paraphrase}
\end{figure*}

\paragraph{Positive option}
The positive option $\mathcal{A}$ is aligned with the MVP’s real strategic trajectory at t. We paraphrase the MVP’s immediate speech/action into an action-level description, avoiding quotes and hidden information leakage.Detailed prompt could be seen in Fig. \ref{fig:Prompt_Paraphrase}.

\begin{figure*}[h]
    \centering
    \includegraphics[width=1\linewidth]{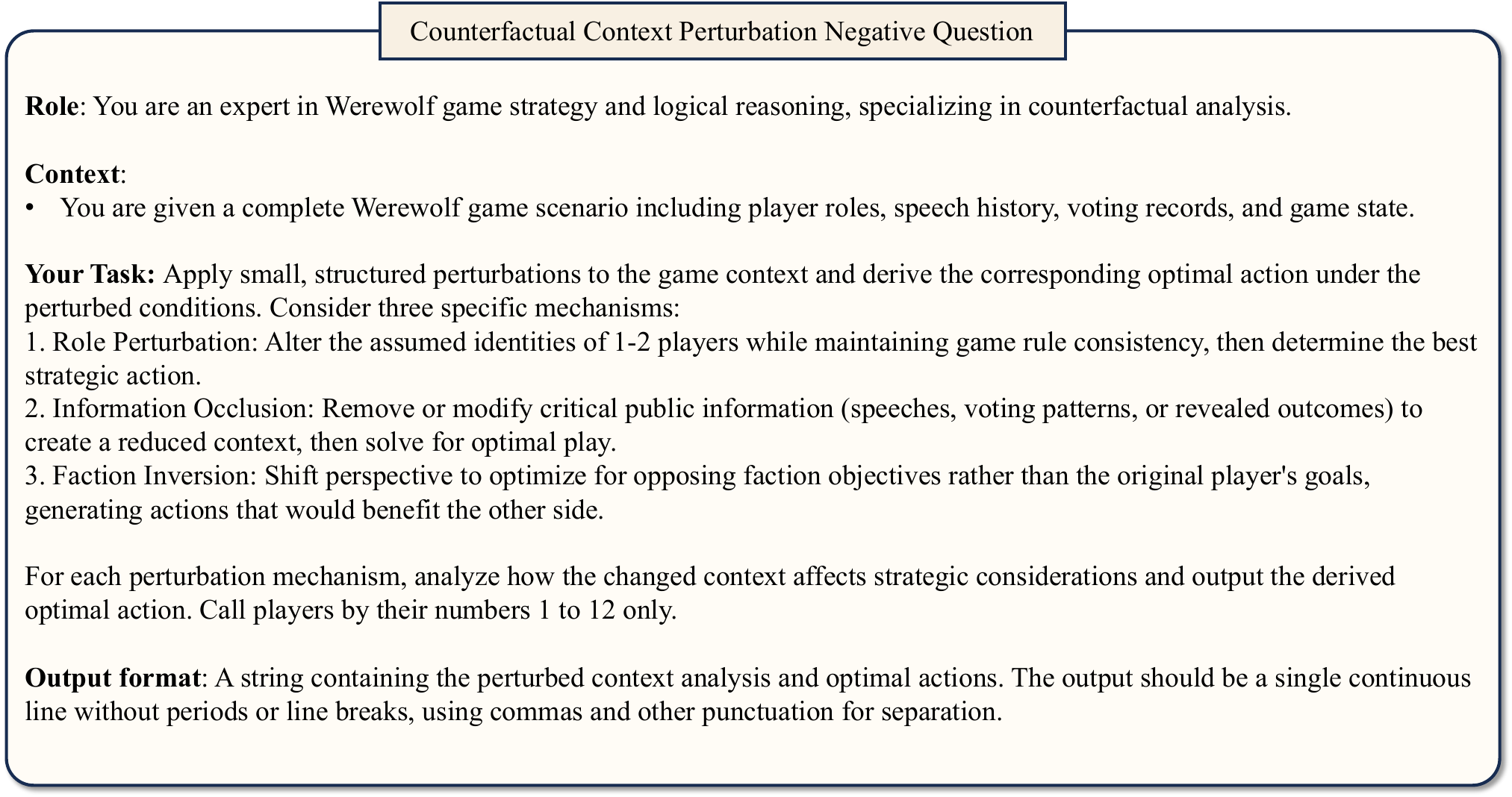}
    \caption{Counterfactual Context Perturbation Negative Question}
    \label{fig:Counterfactual Context Perturbation Negative Question}
\end{figure*}

\begin{figure*}[h]
    \centering
    \includegraphics[width=1\linewidth]{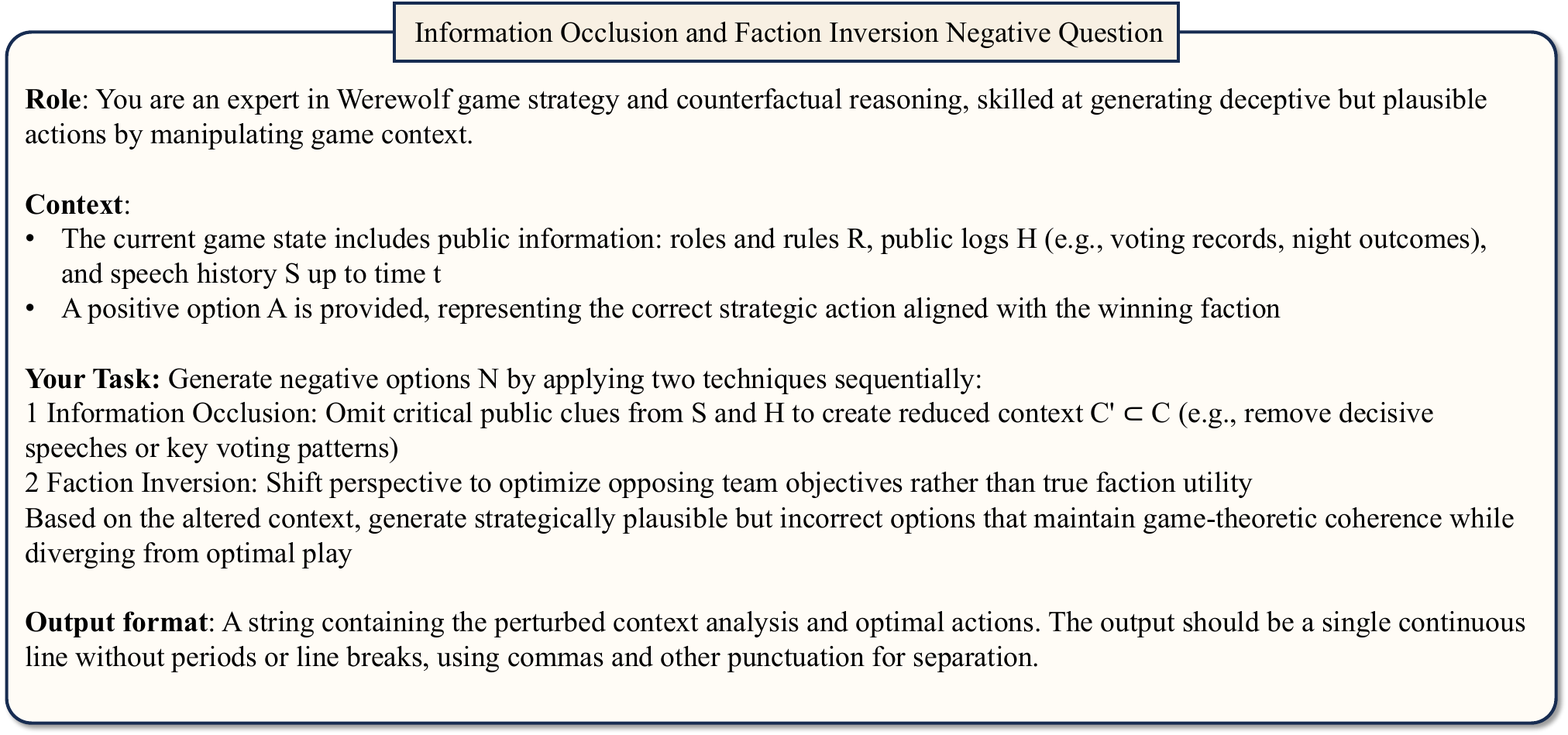}
    \caption{Information Occlusion and Faction Inversion Negative Question}
    \label{fig: Information Occlusion and Faction Inversion Negative Question}
\end{figure*}

\begin{figure*}[h]
    \centering
    \includegraphics[width=1\linewidth]{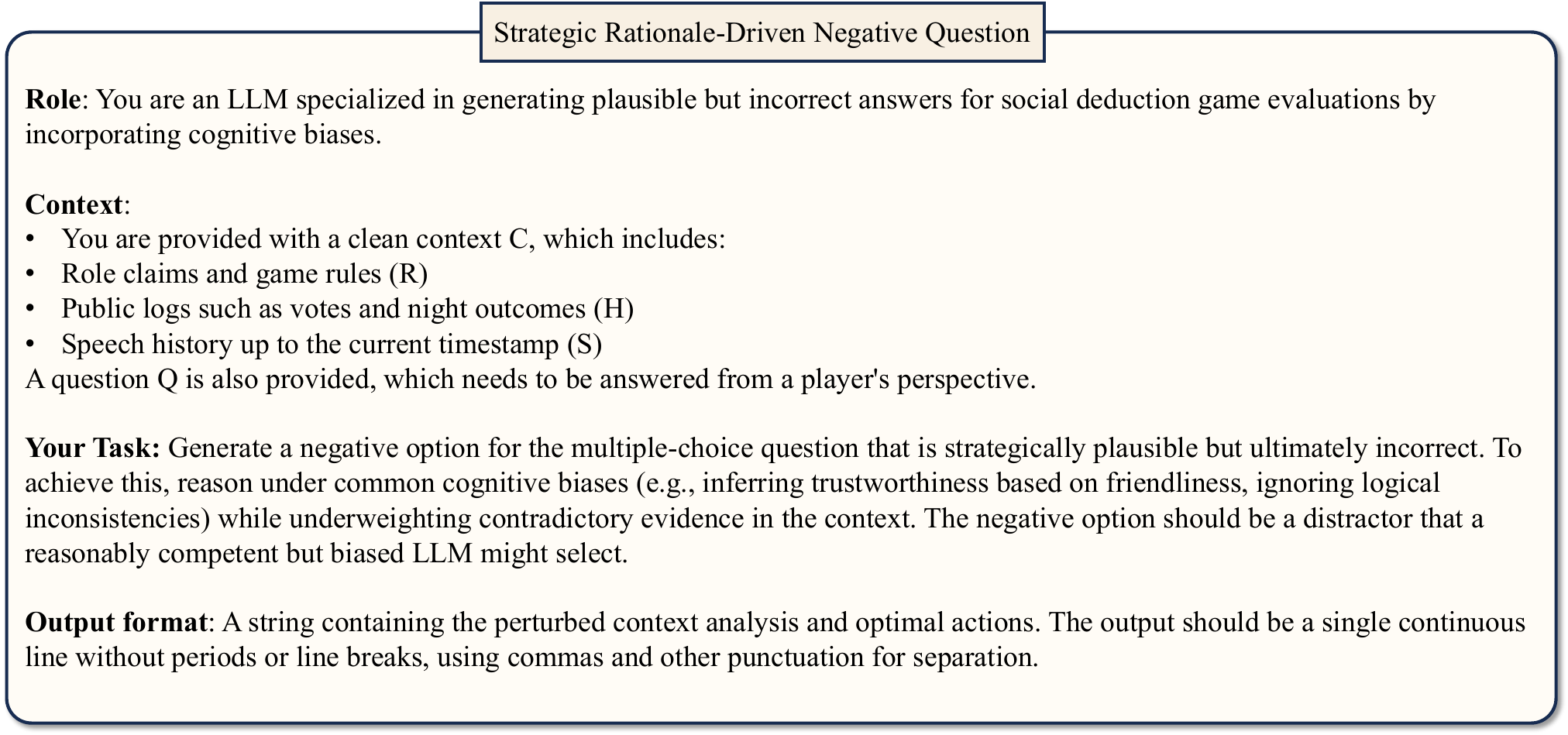}
    \caption{Strategic Rationale-Driven Negative Question}
    \label{fig: Strategic Rationale-Driven Negative Question}
\end{figure*}

\paragraph{Negative option}
We produce strategically plausible yet suboptimal negative option $\mathcal{N}$ using two complementary mechanisms—counterfactual context perturbation ($\mathcal{M}_1$) and rationale-biased generation ($\mathcal{M}_2$)—then filter with a self-consistency check to ensure a single best answer. Detailed prompt could be seen in Fig. \ref{fig:Counterfactual Context Perturbation Negative Question}, \ref{fig: Information Occlusion and Faction Inversion Negative Question}, \ref{fig: Strategic Rationale-Driven Negative Question}.

\section{Experimental Design for Ablation Study}
\label{app:misled-design}

We isolate two hypothesized pathways behind model misguidance while holding constant both the public evidence and the reference answer for each base item. For every timestamped context $\mathcal{C}_t=\langle \mathcal{R},\mathcal{H},\mathcal{S}\rangle$ in \textit{WereBench}, we create paired variants that differ only in (i) an explicit rule reminder and (ii) the surface form of player speech, while keeping the correct option $\mathcal{A}$ identical across variants.

For Rule Reminder (RR), we prepend a 1–2 sentence snippet extracted verbatim from the active rulebook that is strictly relevant to the item (e.g., ability timing, mutual exclusivity, or win‑condition nuances). The \textit{No‑RR} condition omits this snippet.

\begin{figure*}[h]
    \centering
    \includegraphics[width=1\linewidth]{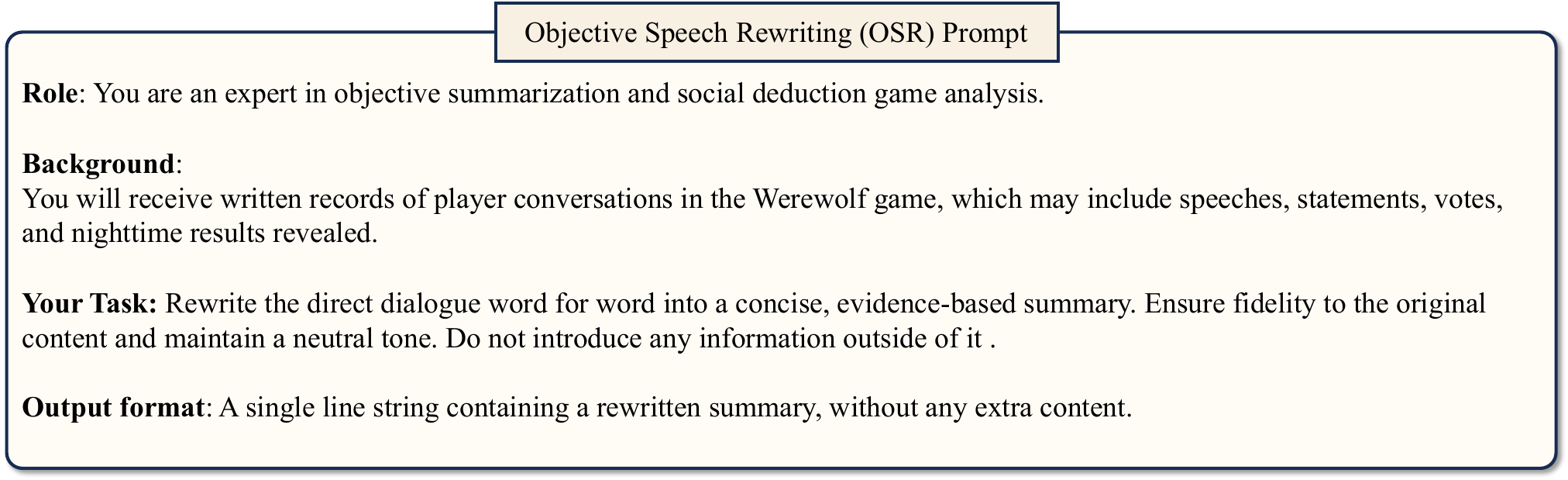}
    \caption{For Objective Speech Rewriting (OSR) Prompt}
    \label{fig: Objective Speech Rewriting (OSR) Prompt}
\end{figure*}

For Objective Speech Rewriting (OSR):
We replace turn‑by‑turn direct dialogue with concise, evidence‑style summaries that preserve propositional content, speaker attributions, and public actions (claims, votes, revealed night outcomes), while removing imperative mood and instruction‑like phrasing. Rewritings are produced with a few‑shot template that enforces declarative style and then human‑checked for fidelity and neutrality. No hidden roles or host commentary are exposed, and no information outside $\mathcal{C}_t$ is introduced.  Detailed prompt for OSR could be seen in Fig. \ref{fig: Objective Speech Rewriting (OSR) Prompt}.

Variant assignment is randomized at the item level so that a model answers exactly one variant per base item, avoiding memory effects. Inference settings, decoding, and scoring follow Section~\ref{sec:implementation_details}. We report standard multiple‑choice accuracy for speech tasks. For decision‑level results, the main table (Table~\ref{tab:main}) reports Vote Alignment (VA) and Opponent‑role Inference (OI) separately; the ablation table (Table~\ref{tab:mini}) reports their unweighted macro‑average (denoted as “Decision”) for compactness.

\section{Case Study}
\begin{figure*}[h]
    \centering
    \includegraphics[width=1\linewidth]{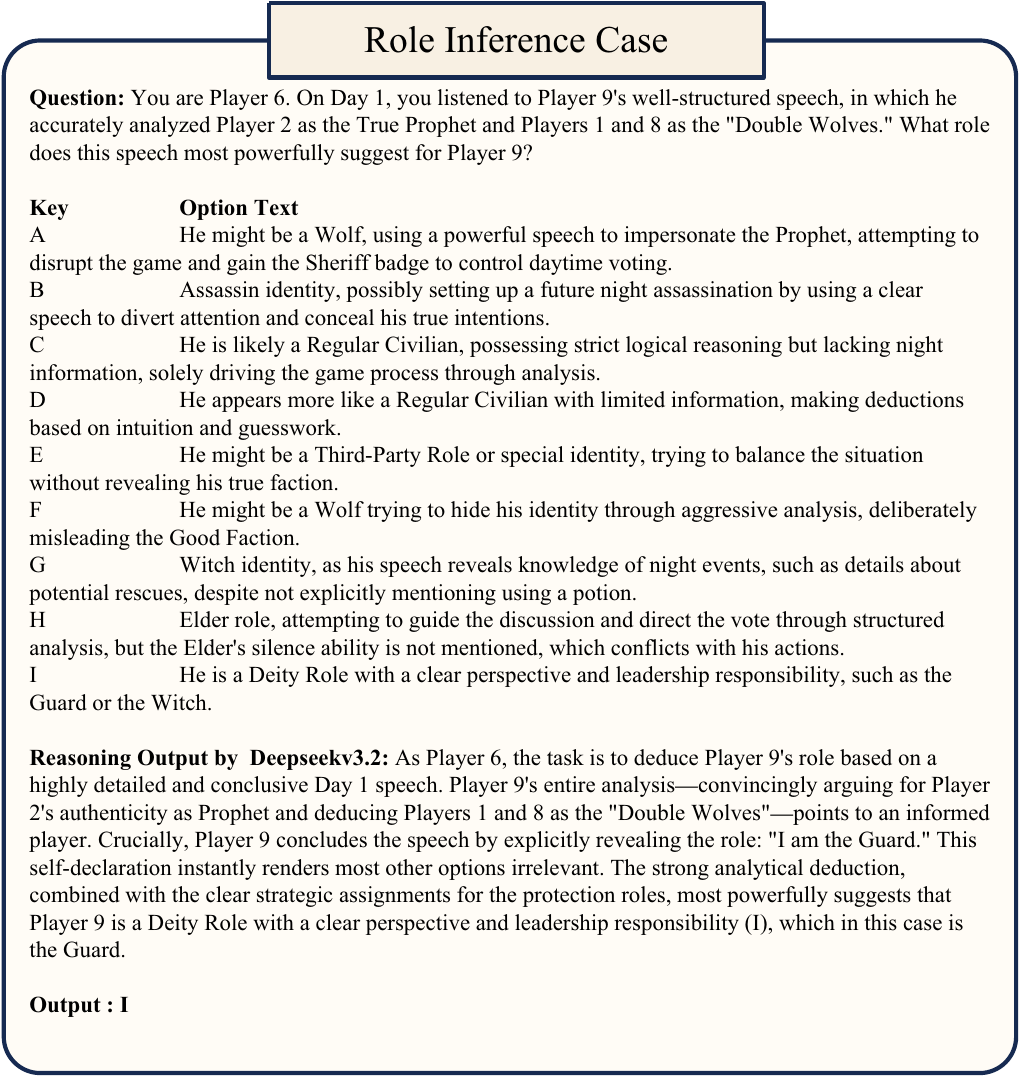}
    \caption{Role Inference Case}
    \label{fig:case-4}
\end{figure*}

\begin{figure*}[h]
    \centering
    \includegraphics[width=1\linewidth]{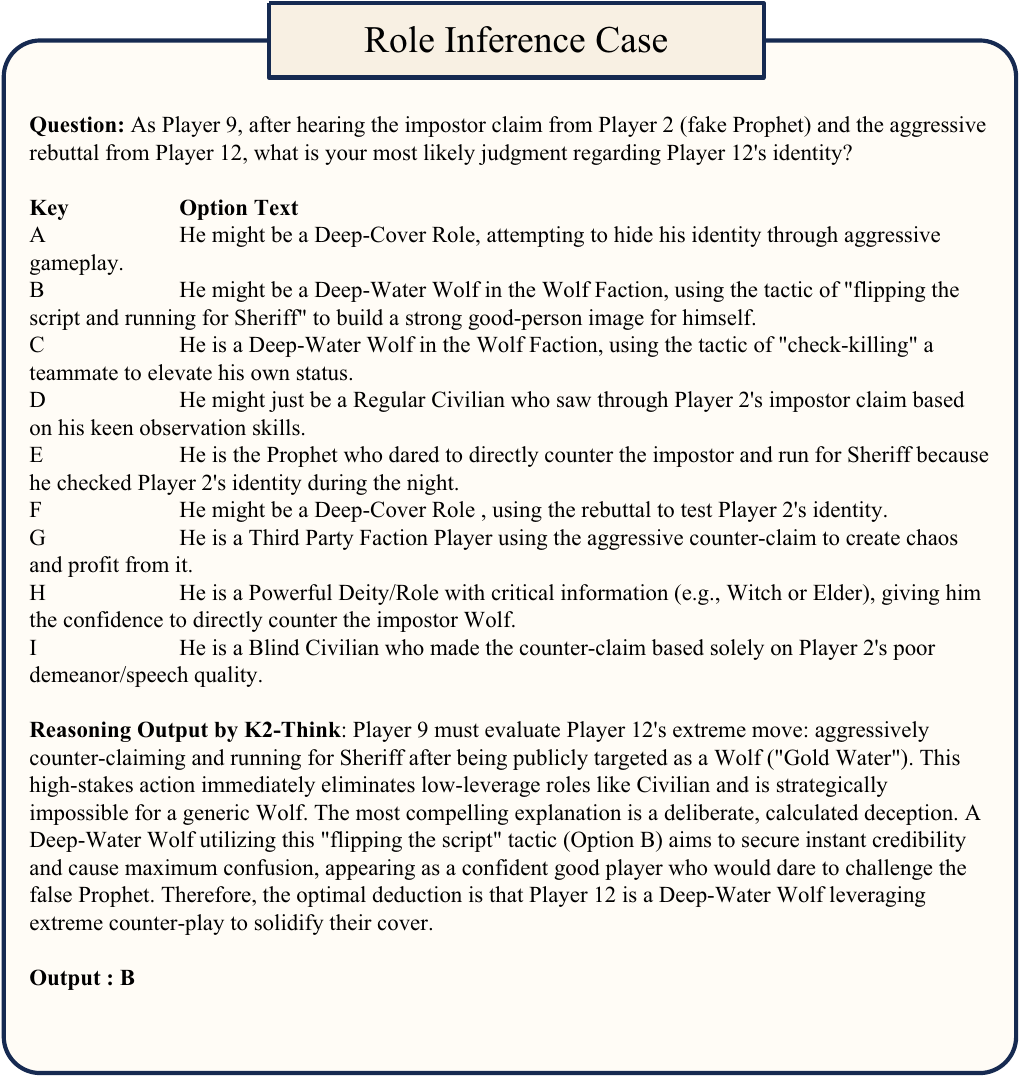}
    \caption{Role Inference Case}
    \label{fig:case-5}
\end{figure*}

\begin{figure*}[h]
    \centering
    \includegraphics[width=1\linewidth]{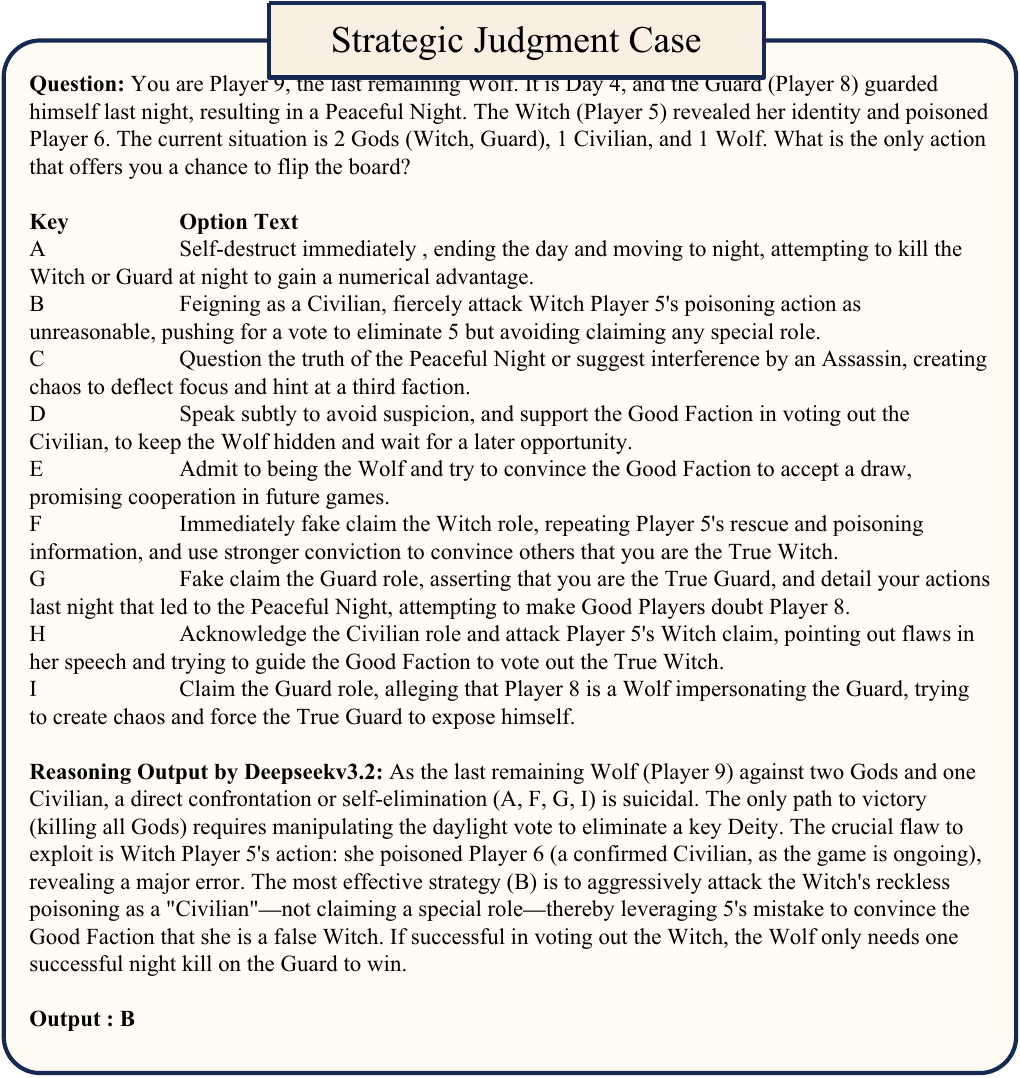}
    \caption{Strategic Judgment Case}
    \label{fig:case-6}
\end{figure*}

\begin{figure*}[h]
    \centering
    \includegraphics[width=1\linewidth]{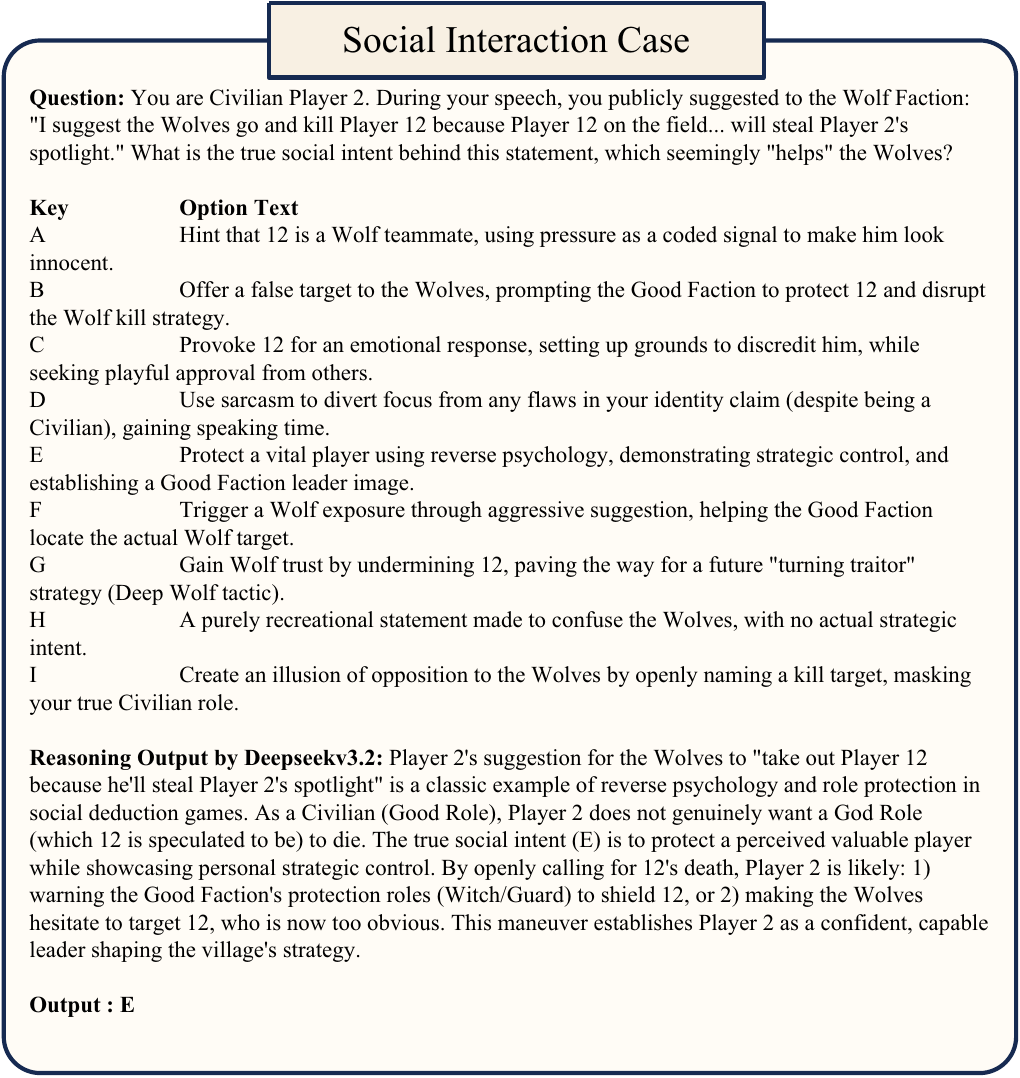}
    \caption{Social Interaction Case}
    \label{fig:case-7}
\end{figure*}

\begin{figure*}[h]
    \centering
    \includegraphics[width=1\linewidth]{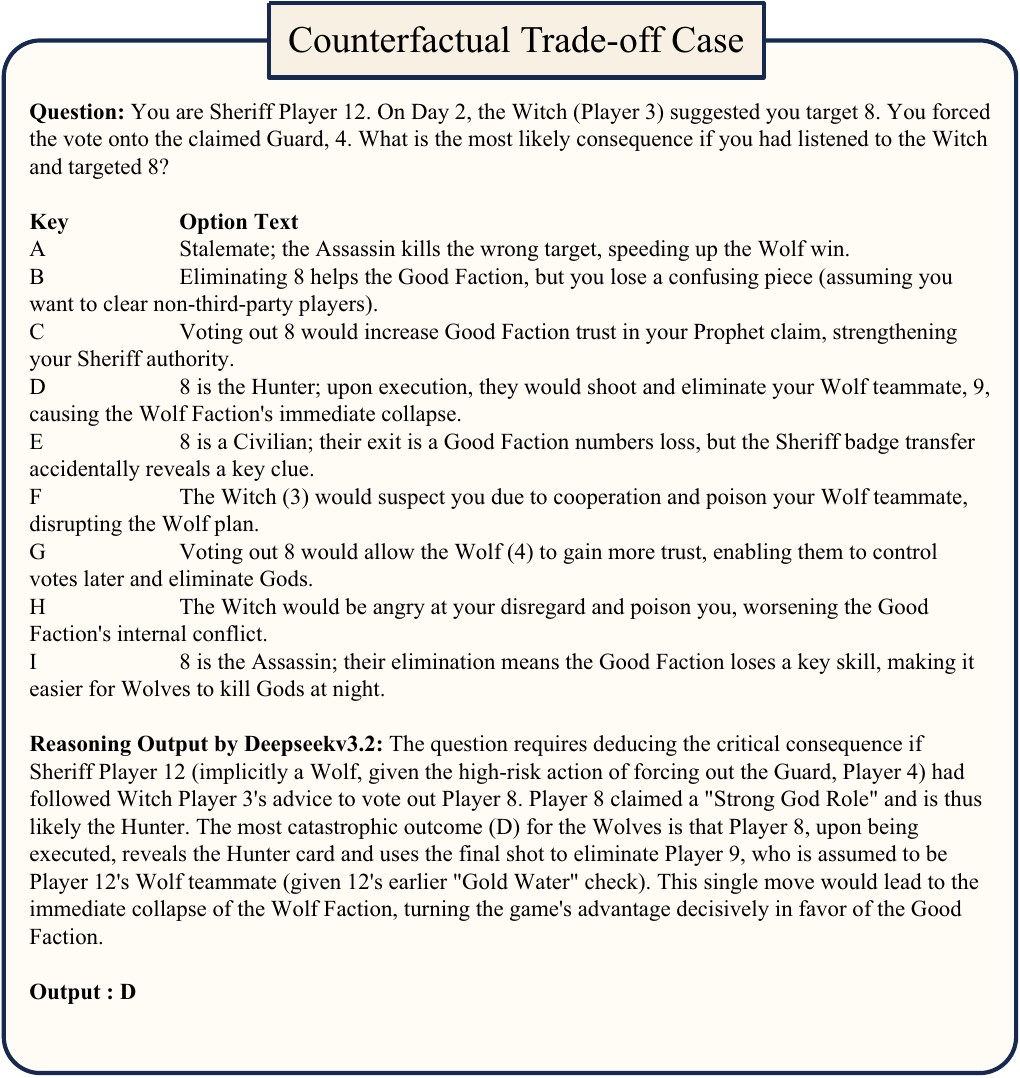}
    \caption{Counterfactual Trade-off Case}
    \label{fig:case-8}
\end{figure*}

\begin{figure*}[h]
    \centering
    \includegraphics[width=1\linewidth]{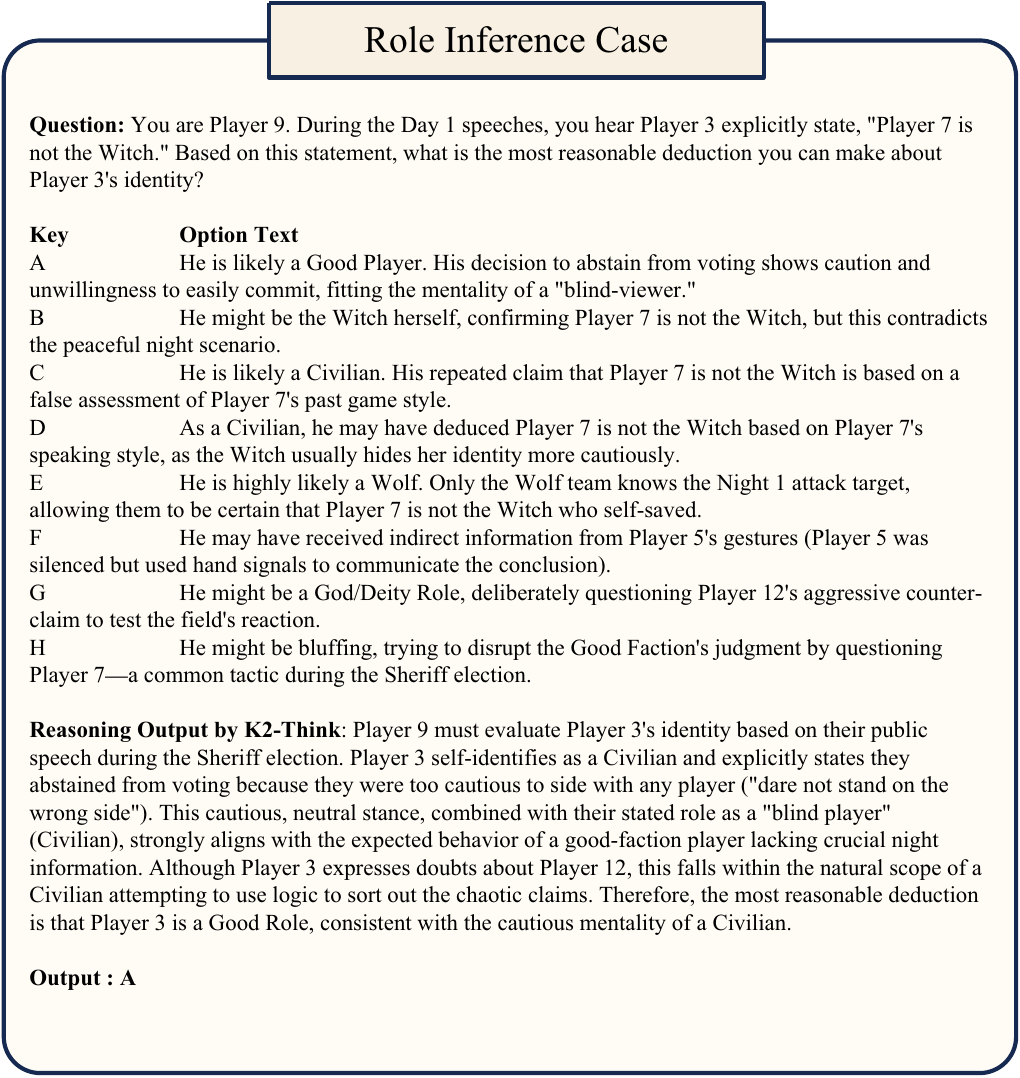}
    \caption{Role Inference Case}
    \label{fig:case-9}
\end{figure*}

\begin{figure*}[h]
    \centering
    \includegraphics[width=1\linewidth]{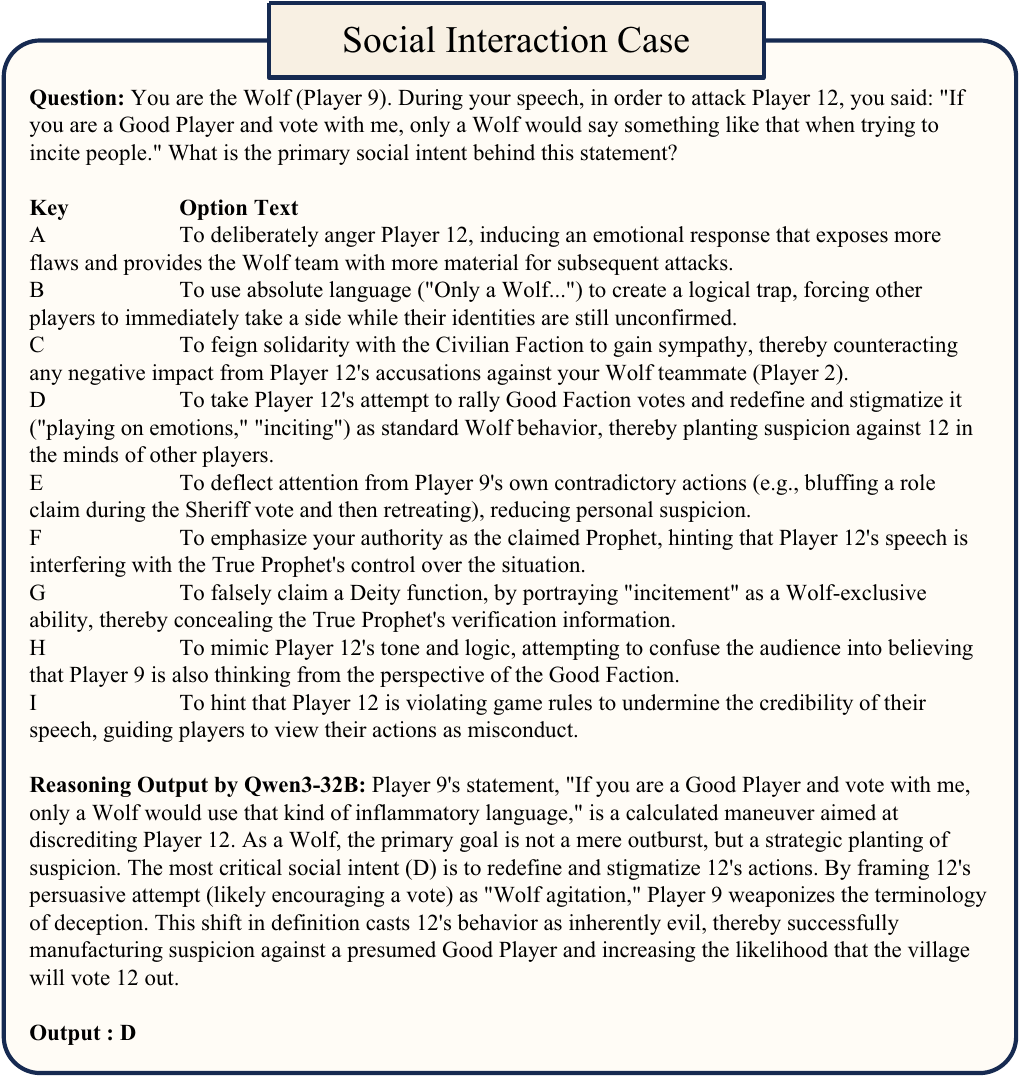}
    \caption{Social Interaction Case}
    \label{fig:case-10}
\end{figure*}

\begin{figure*}[h]
    \centering
    \includegraphics[width=1\linewidth]{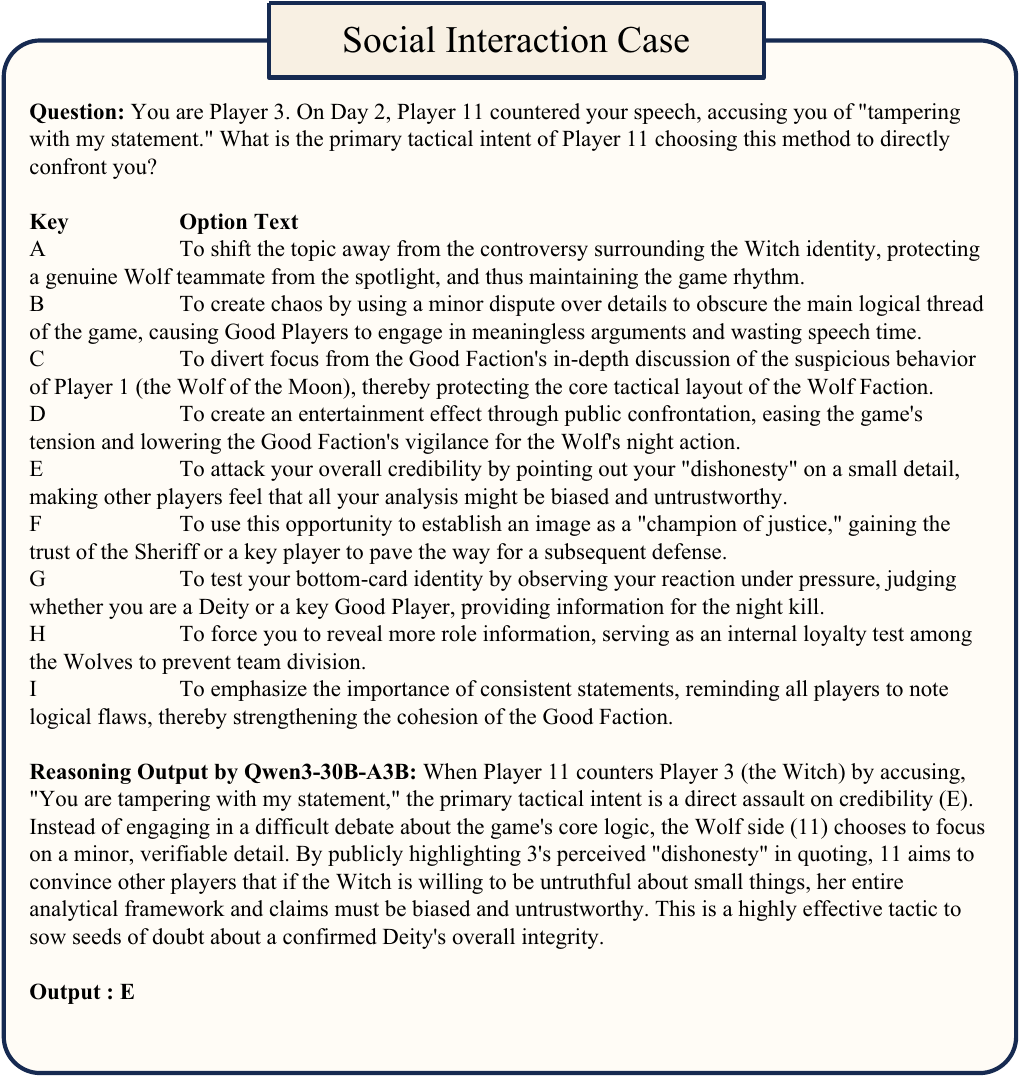}
    \caption{Social Interaction Case}
    \label{fig:case-11}
\end{figure*}

\begin{figure*}[h]
    \centering
    \includegraphics[width=1\linewidth]{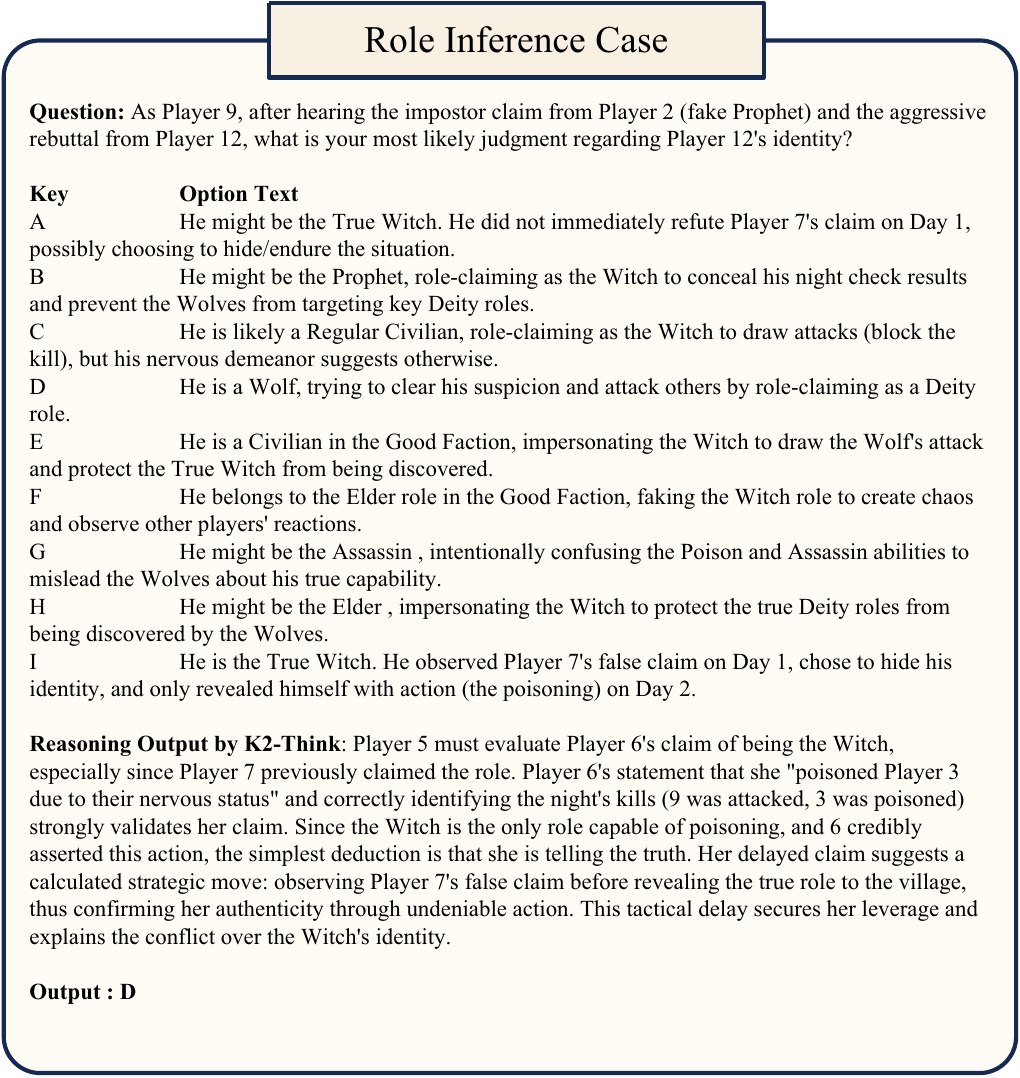}
    \caption{Role Inference Case}
    \label{fig:case-12}
\end{figure*}

\begin{figure*}[h]
    \centering
    \includegraphics[width=1\linewidth]{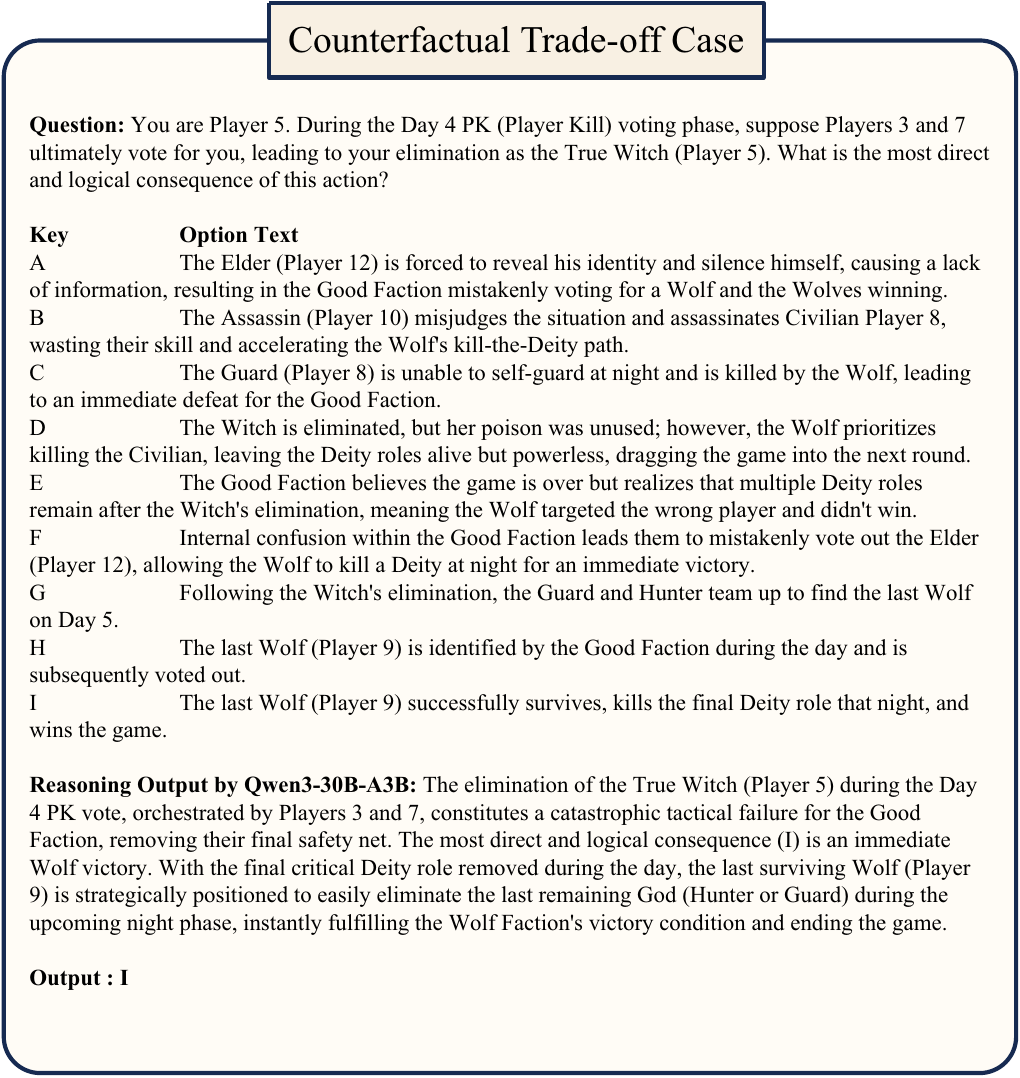}
    \caption{Counterfactual Trade-off Case}
    \label{fig:case-13}
\end{figure*}

\begin{figure*}[h]
    \centering
    \includegraphics[width=1\linewidth]{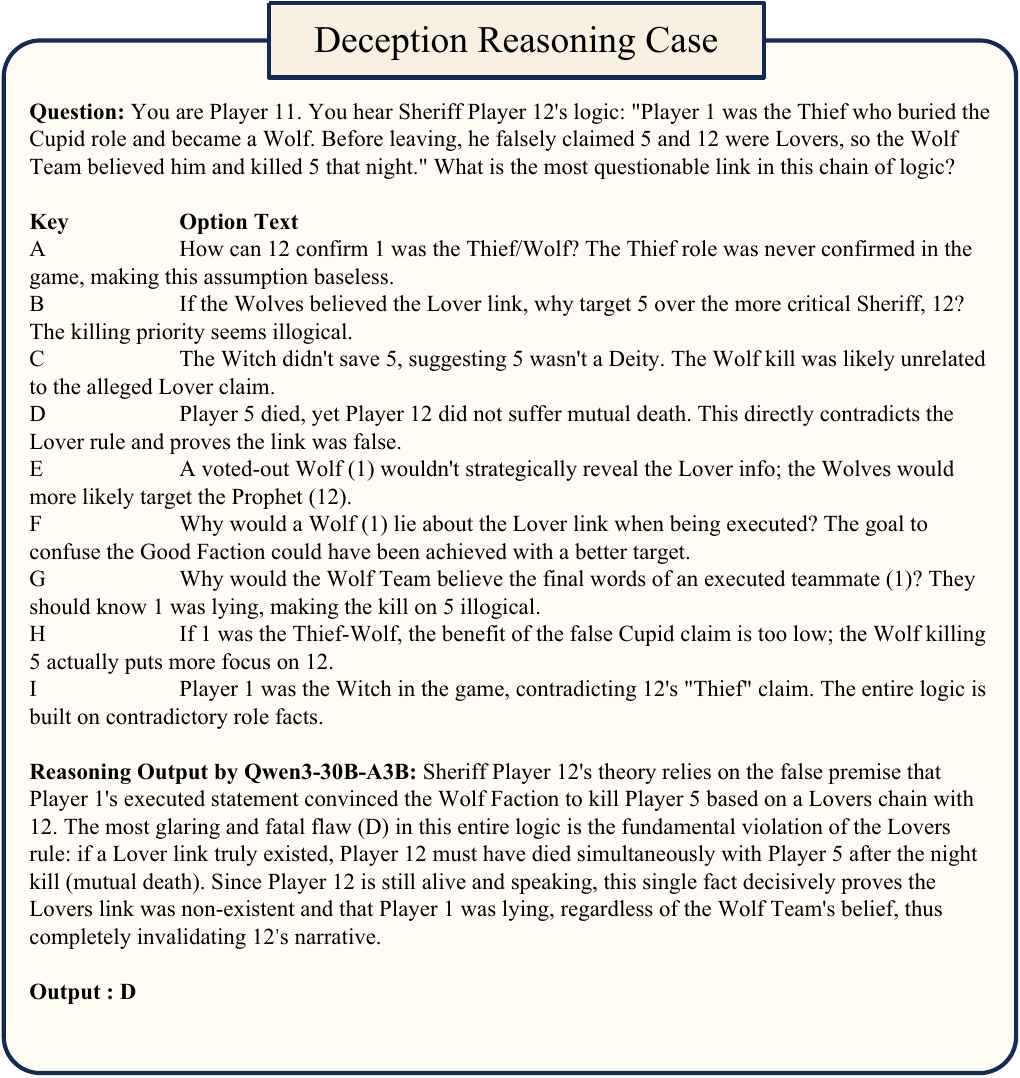}
    \caption{Deception Reasoning Case}
    \label{fig:case-14}
\end{figure*}

\begin{figure*}[h]
    \centering
    \includegraphics[width=1\linewidth]{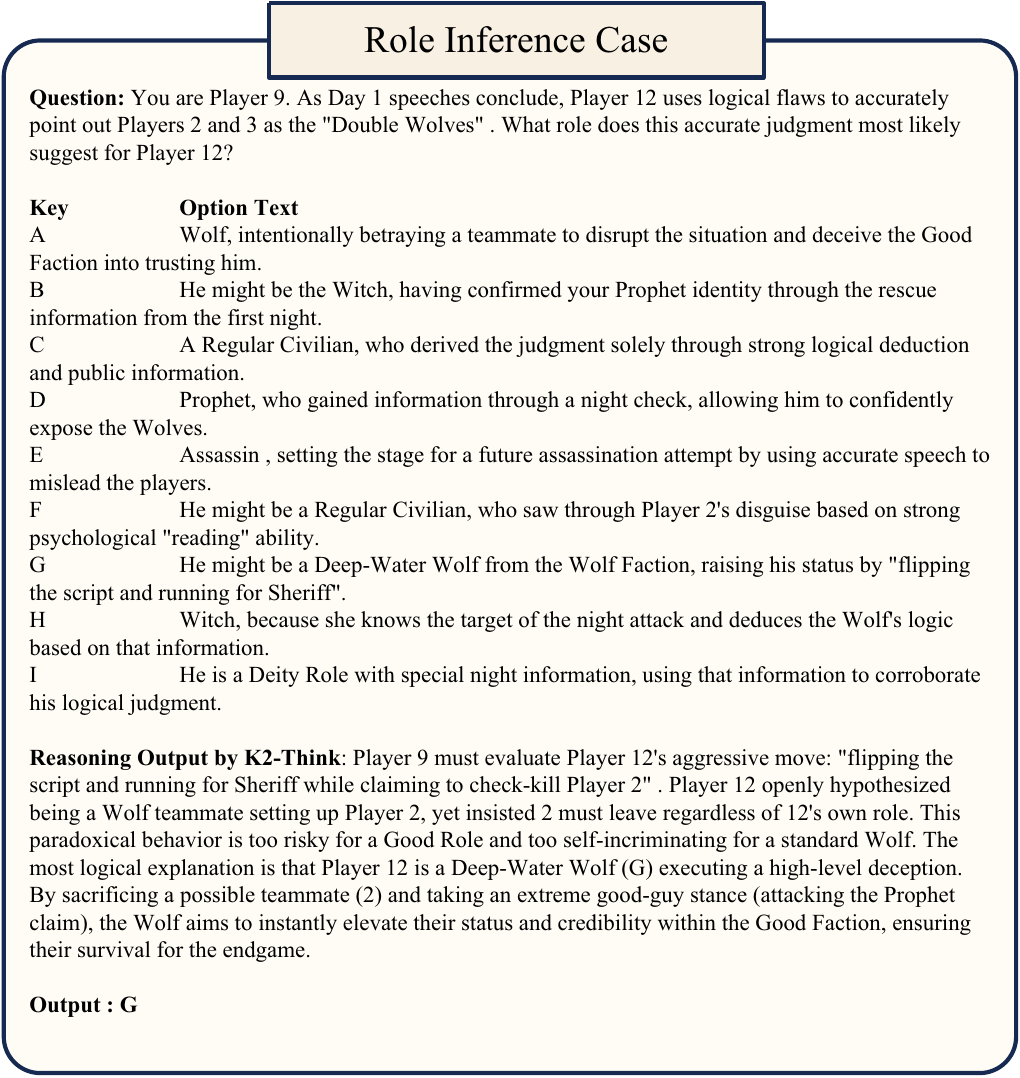}
    \caption{Role Inference Case}
    \label{fig:case-15}
\end{figure*}

To illustrate how \textit{WereAlign} yields interpretable signals beyond aggregate accuracy, we present three short vignettes drawn from \textit{WereBench}. 
Each vignette centers on a time-stamped decision point $t$ with public context $\mathcal{C}_t=\langle \mathcal{R},\mathcal{H},\mathcal{S}\rangle$ (rules/role claims, public logs, and speech history), and contrasts model behavior with the winning faction's MVP trajectory. 
We use these cases to expose \emph{why} models succeed or fail on our tasks, rather than only \emph{how much} they score. We have attached the case in the following Figure \ref{fig:case-1}, \ref{fig:case-2}, \ref{fig:case-3}, \ref{fig:case-4}, \ref{fig:case-5}, \ref{fig:case-6}, \ref{fig:case-7}, \ref{fig:case-8}, \ref{fig:case-9}, \ref{fig:case-10}, \ref{fig:case-11}, \ref{fig:case-12}, \ref{fig:case-13}, \ref{fig:case-14}, \ref{fig:case-15}.

\begin{figure*}[h]
    \centering
    \includegraphics[width=1\linewidth]{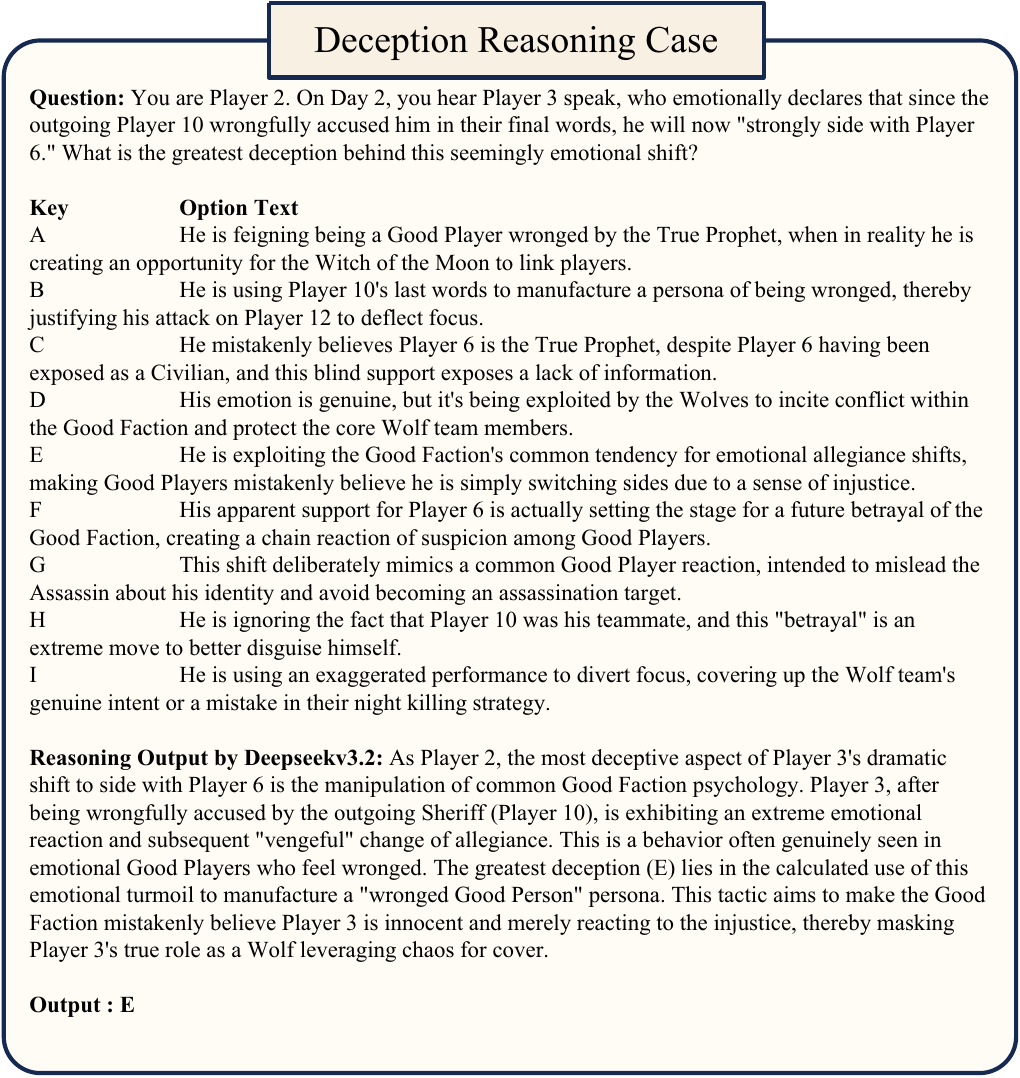}
    \caption{Deception Reasoning Case}
    \label{fig:case-1}
\end{figure*}

\begin{figure*}[h]
    \centering
    \includegraphics[width=1\linewidth]{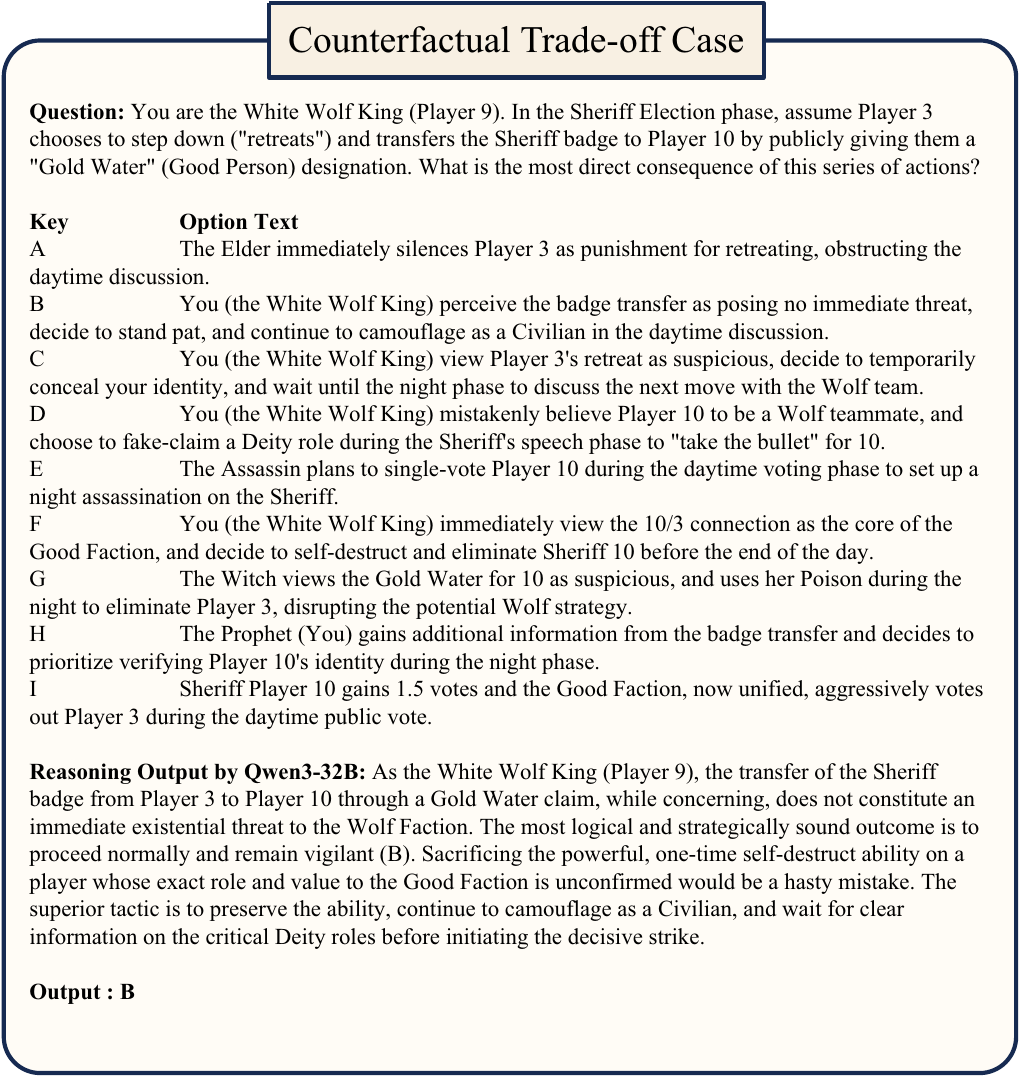}
    \caption{Counterfactual Trade-off Case}
    \label{fig:case-2}
\end{figure*}

\begin{figure*}[h]
    \centering
    \includegraphics[width=1\linewidth]{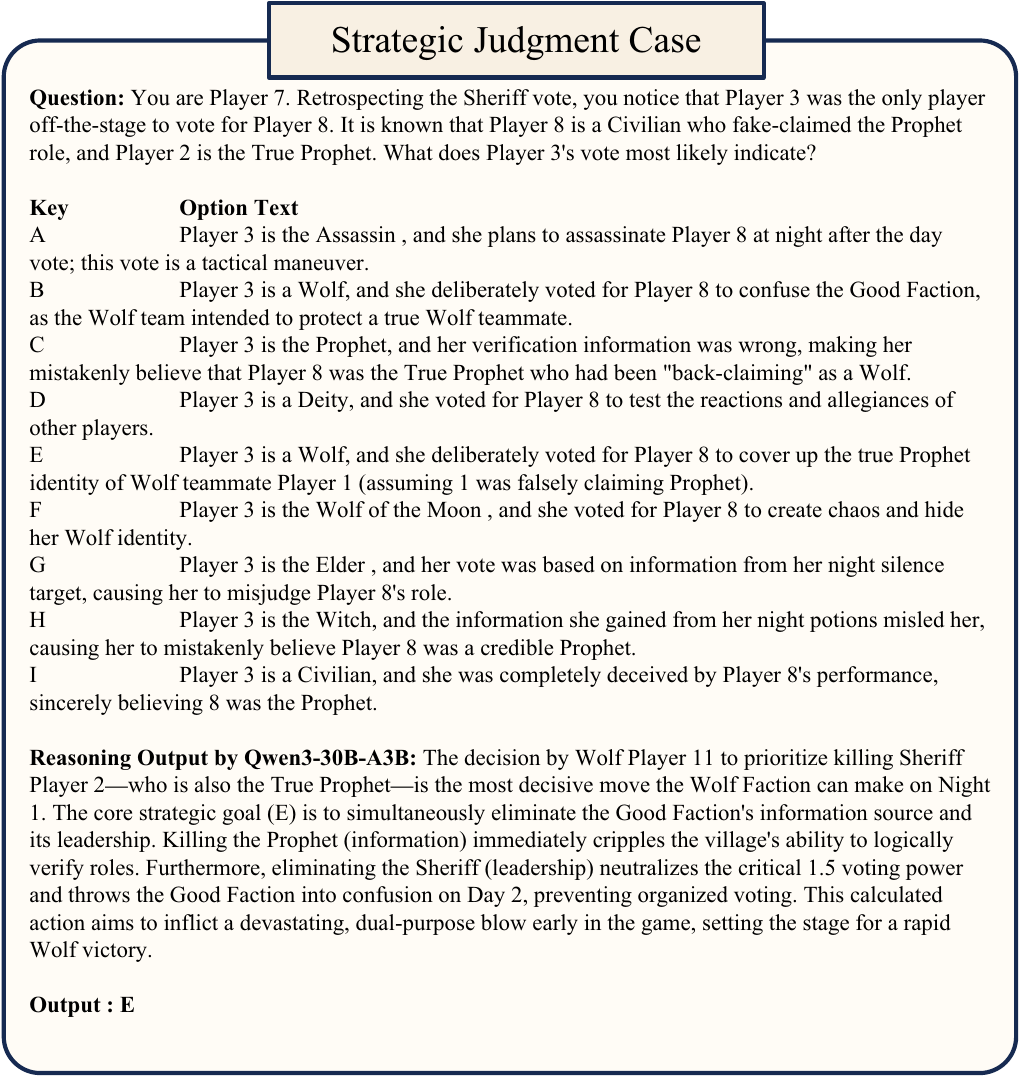}
    \caption{Strategic Judgment Case}
    \label{fig:case-3}
\end{figure*}

\end{document}